\pdfoutput=1

\documentclass[11pt]{article}

\usepackage[]{EMNLP2023}

\usepackage{times}
\usepackage{latexsym}

\usepackage[T1]{fontenc}

\usepackage[utf8]{inputenc}

\usepackage{microtype}

\usepackage{inconsolata}

\usepackage{array}
\usepackage{makecell}

\usepackage{graphicx,multirow}
\usepackage{url}
\usepackage{hyperref}
\newcommand\blfootnote[1]{%
  \begingroup
  \renewcommand\thefootnote{}\footnote{#1}%
  \addtocounter{footnote}{-1}%
  \endgroup
}

\title{Multimodal Automated  Fact-Checking: A Survey}

\author{Mubashara Akhtar\textsuperscript{1,*}, Michael Schlichtkrull\textsuperscript{2}, Zhijiang Guo\textsuperscript{2}, Oana Cocarascu\textsuperscript{1}, \\ {\bf Elena Simperl\textsuperscript{1}} \and {\bf Andreas Vlachos\textsuperscript{2}} \\[5pt]
        \textsuperscript{1}Department of Informatics, King's College London\\ 
        \textsuperscript{2}Department of Computer Science and Technology, University of Cambridge\\[5pt]
        \texttt{\{mubashara.akhtar,oana.cocarascu,elena.simperl\}@kcl.ac.uk}\\ \texttt{\{mss84,zg283,av308\}@cam.ac.uk}}

\begin{document}
\maketitle
\begin{abstract}

Misinformation 
is often conveyed 
in multiple modalities, e.g.\ a miscaptioned image. 
Multimodal misinformation 
is perceived as more credible by humans, and spreads faster 
than its text-only counterparts. %
While an increasing body of research investigates automated fact-checking (AFC), previous surveys mostly focus on text. 
In this survey, we conceptualise a framework for AFC including subtasks unique to multimodal misinformation.
Furthermore, we 
discuss related terms
used in different communities 
and map them to our framework.
We focus on four modalities prevalent in real-world fact-checking: text, image, audio, and video. We survey benchmarks and models, and discuss limitations and promising directions for future research.\blfootnote{\textsuperscript{*} This work was partially done during Mubashara's research visit at Cambridge.}




\end{abstract}
    
\section{Introduction}

Motivated by the challenges presented by misinformation in the modern media ecosystem, previous research has commonly modelled automated fact-checking (AFC) as a pipeline consisting of different stages, surveyed in a variety of axes~\citep{thorne-vlachos-2018-automated, kotonya-toni-2020-explainable, DBLP:journals/llc/ZengAZ21, NakovCHAEBPSM21, guo-etal-2022-survey}.
%
%
%
However, these surveys focus on a single modality, text.
This is different to real-world misinformation that often occurs via several modalities. 
\blfootnote{\textsuperscript{1}\url{https://www.bbc.com/news/world-us-canada-65069316}}


\begin{figure}
\centering
\includegraphics[scale=0.26]{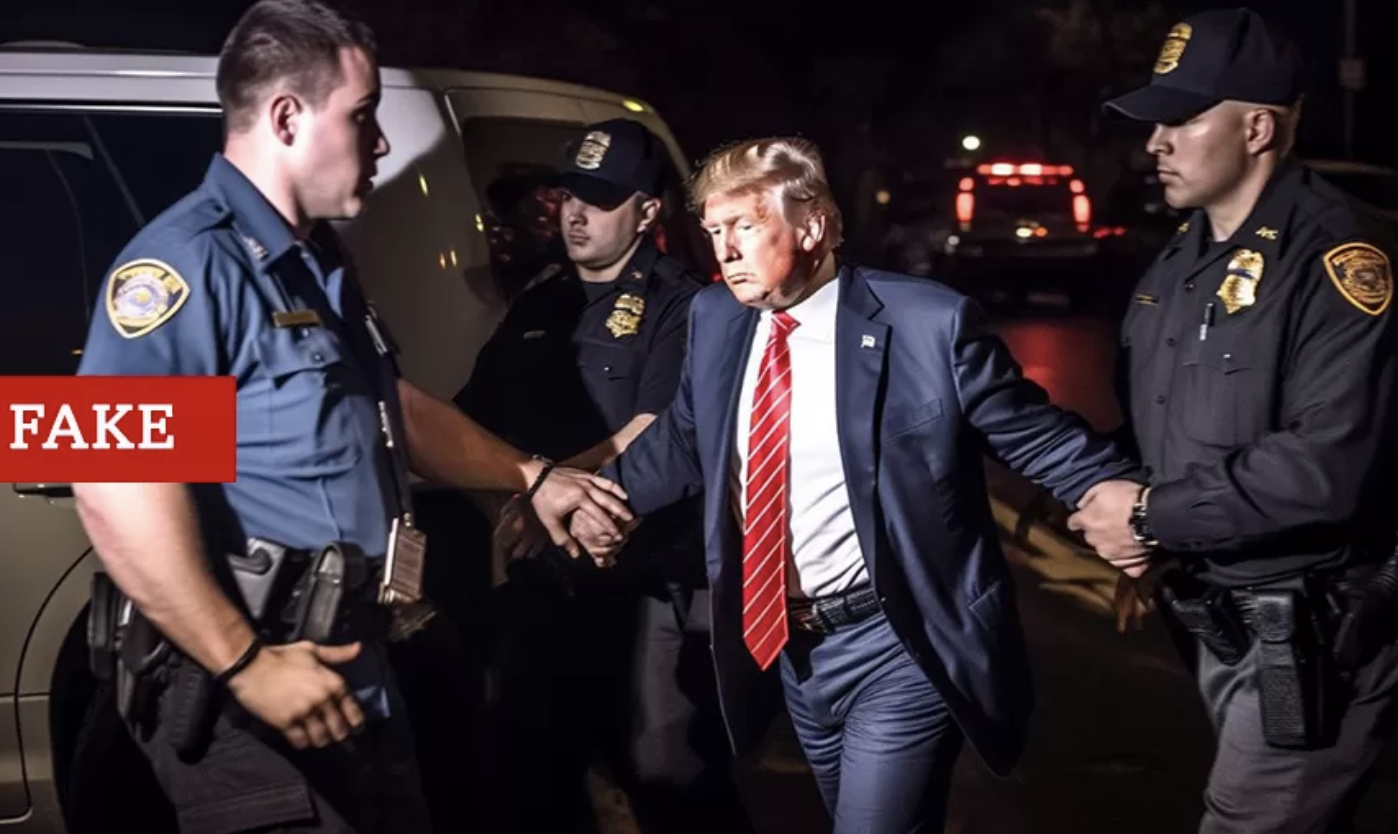}
\caption{\label{fig:deepfake} Manipulated image depicting arrest of former US president Donald Trump (source: BBC\footnote{https://www.bbc.com/news/world-us-canada-65069316}).
}
\end{figure}

In 
AFC, the term \emph{multimodal} has been used to refer to cases where the claim and/or evidence are expressed through different or multiple modalities~\citep{ hameleers2020picture, alam-etal-2022-survey, biamby-etal-2022-twitter}. 
Examples of 
multimodal misinformation
include: 
$(i)$ claims about digitally manipulated content~\citep{DBLP:conf/cvpr/AgarwalFGHN019, Andreas18} such as photos depicting former US president Trump's arrest (Figure \ref{fig:deepfake});
$(ii)$ combining content from different modalities and contexts, e.g.\ using 
video footage in a misleading context~\citep{DBLP:journals/corr/abs-2101-06278, biamby-etal-2022-twitter, Abdelnabi_2022_CVPR};
$(iii)$ embedding a claim in another modality, e.g.\ a meme, an image with embedded text~\citep{qu2022}, with notable real-world examples including a 
Brexit Vote Leave poster\footnote{\url{https://www.itv.com/news/2019-01-18/boris-johnson-under-attack-over-turkey-claim/}} and TikTok videos with COVID misinformation~\citep{ShangKZ021};
$(iv)$ verifying a claim with evidence from a different modality than the input claim, e.g.\ verifying images against text~\citep{Shao_2023_CVPR}, audio against textual metadata~\citep{DBLP:conf/asru/KopevAKN19}, and text against images~\citep{akhtar-etal-2023-reading}.

Fact-checking multimodal misinformation is important for a number of reasons.
First, multimodal content is perceived as more credible compared to text containing a similar claim~\citep{newman2012nonprobative}. 
For example, previous research shows that visual content exhibits a ``photo truthiness''-effect~\citep{inbook}, biasing readers to believe a claim is true.
Second, multimodal content spreads faster and has a higher engagement than text-only posts~\citep{li2020picture}. 
%
Third, with recent advances in generative machine learning models~\citep{Rombach_2022_CVPR}, the generation 
of multimodal misinformation has been simplified.

\begin{table}
\begin{center}
\scalebox{0.95}{
\begin{tabular}{lr}\hline
Claim Modality       &  Percentage\\ \hline
Image                 &  20.07\% \\
Video                &   8.06\% \\
Audio &  0.55\% \\ \hline
Total & 28.68\% \\ \hline
\end{tabular}}
\caption{Non-textual modalities present and/or used in addition to text in our manually annotated snapshot of real-world claims from the Google ClaimReview API.}
\label{table:averitec_claim_count}
\end{center}
\end{table}

To validate the importance of multimodal fact-checking, we manually annotated 9,255 claims from the AVeriTeC dataset~\citep{schlichtkrullGV2023}, which were collected with
the Google FactCheck ClaimReview API\footnote{\url{https://toolbox.google.com/factcheck/apis}}. For each claim, we identified the modalities present in it and evidence strategies (e.g.\ identification of geolocation) used for fact-checking. We find that more than $2,600$ (28.68\%) claims either contain multimodal data or require multimodal reasoning for verification, with 20.07\% involving images, 8.06\% videos, and 0.55\% audios (see Table~\ref{table:averitec_claim_count}).\footnote{Annotations at \url{http://github.com/MichSchli/AVeriTeC}.} These claims 
can neither be fact-checked by a text-only model, nor by a model with no text capabilities.





In this survey, we introduce a three-stage framework for multimodal automated fact-checking
: claim detection and extraction, evidence retrieval, and verdict prediction encompassing veracity,  manipulation and out-of-context classification, as well as justification production. 
The input and output data of each stage 
can have different or multiple modalities.
For each stage, we discuss related terms and definitions developed in different research communities.
In contrast to previous surveys on multimodal fact-checking that focus on individual subtasks ~\citep{DBLP:journals/corr/abs-2003-05096, alam-etal-2022-survey,DBLP:journals/corr/abs-2203-13883}, we consider all subtasks
surveying benchmarks and modeling approaches for them. 

We focus on four prevalent modalities of real-world fact-checking identified in our annotations: text, image, audio, and video. 
While tables and knowledge graphs are increasingly used as evidence for benchmarks~\citep{DBLP:conf/iclr/ChenWCZWLZW20, DBLP:conf/nips/AlyGST00CM21, akhtar-etal-2022-pubhealthtab}, they have been 
covered in previous surveys~\citep{thorne-vlachos-2018-automated,  DBLP:journals/llc/ZengAZ21, guo-etal-2022-survey}. 
Finally, we discuss 
the extent to which current approaches to AFC work for multimodal data,
and promising directions for further research 
(Section~\ref{sec:challenges}).
We accompany the survey with a repository,\footnote{\url{https://github.com/Cartus/Automated-Fact-Checking-Resources}} which lists the resources mentioned in our survey.

\section{Task Formulation}
\label{sec:taskformulation}

\begin{figure*}
\centering
\includegraphics[scale=0.8]{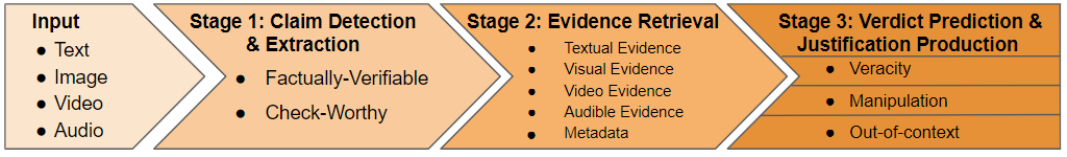}
\caption{\label{fig:overviewtask} Multimodal fact-checking pipeline.}
\end{figure*}

This section introduces a conceptualisation of multimodal AFC as a three-stage process, including claim detection and extraction, evidence retrieval, and production of verdicts and justifications for various types of misinformation  (Figure~\ref{fig:overviewtask}). Compared to the text-only pipeline presented in \citet{guo-etal-2022-survey}, our framework extends their first stage with a claim \textit{extraction} stage, and generalises their third stage to cover tasks that fall under multimodal AFC, thus accounting for its particular challenges. 

\paragraph{Terminology.} A number of works \citep{DBLP:conf/icwsm/SinghalSK22, fung-etal-2021-infosurgeon} use the term \emph{multimedia}, which is also more common in public discussions instead of \emph{multimodal}~\citep{LAUER2009225}. However in 
in this survey we adopt the latter, following other surveys
that use multimodal data~\citep{DBLP:journals/corr/LiangZM22, DBLP:journals/access/GuoWW19}.
Adopting the terminology of previous surveys~\citep{thorne-vlachos-2018-automated, alam-etal-2022-survey} and following advice from institutions such as the UNO~\citep{ireton2018journalism}, we avoid \emph{multimodal fake news}~\citep{MEEL202123, DBLP:conf/fps/AmriSA21, PatwaMSBCRDCSEA22} due to the term's ambiguous use. 
\paragraph{Stage $1$: Claim Detection and Extraction.}
\label{ssec:claimdetection}
The first pipeline stage
aims to find \textit{checkable} (i.e.\ factually-verifiable) and \textit{check-worthy} (i.e.\ important factual claims~\citep{DBLP:conf/cikm/HassanLT15}) claims.
Debunking a typical claim and writing the fact-checking article takes approximately one day for a human fact-checker~\citep{hassan2015quest}. This stage aims 
to focus the AFC process on claims which are verifiable and most impactful. 
%
Multimodal claims can be diverse and include: $(1)$ 
a written claim embedded in another modality~\citep{PrabhakarGNG21, DBLP:conf/misdoom/MarosAV21} such as an image or a spoken claim in an audio or video; 
$(2)$ a claim that a piece of content is authentic, e.g.\ that a video footage is from a specific geographic location~\citep{ZhangSGLLZA018, Heller18}; 
$(3)$ a claim for which the evidence is manipulated to support it, e.g.\ through lip-syncing~\citep{Andreas18}. 
While in some cases the claim is clearly specified (e.g. in form of a headline), in often multiple modalities are required to understand and extract a claim at this stage.
Simply \textit{detecting} potentially misleading content is often not enough -- it is necessary to \textit{extract} the claim before fact-checking it in the subsequent stages.
For example, 
detecting text in images or videos and understanding it given the context~\citep{DBLP:journals/corr/abs-2205-12617} or verifying audios by transcribing and extracting claims~\citep{DBLP:conf/misdoom/MarosAV21}.
\paragraph{Stage $2$: Evidence Retrieval.}
Similarly to fact-checking with text, multimodal fact-checking often
relies on evidence to make judgments, similar to the process followed by human fact-checkers 
~\citep{silverman2013verification, NakovCHAEBPSM21}. 
Two main approaches have been used in the past: 
$(i)$ using the claim to-be-checked as evidence itself, e.g.\ to 
detect manipulation~\citep{DBLP:conf/icdm/QiCYGL19,DBLP:conf/icpr/BonettiniCMBBT20};
this can be seen as the multimodal version of evidence-free fact-checking of text claims by checking logical fallacies in the text~\citep{jin-etal-2022-logical}, and 
$(ii)$ retrieving additional evidence 
~\citep{Abdelnabi_2022_CVPR}.
In multimodal fact-checking, the evidence modality can be different from the claim modality. For example, to retrieve evidence for image or audio fact-checking, previous works have also used text 
e.g.\ metadata, social media comments, or captions~\citep{Gupta13, HuhLOE18, Muller-BudackTD20, DBLP:conf/asru/KopevAKN19}. 

\paragraph{Stage $3$: Verdict Prediction and Justification Production.}
Following the fact-checking process of professional fact-checkers, the final stage comprises verdict prediction and the production of justification that explains the fact-check to humans~\citep{graves2018understanding}. 
Verdict prediction is decomposed into three tasks considering
prevalent multimodal misinformation types:
manipulation, using 
content out-of-context, and veracity classification.

\paragraph{Stage $3.1$: Manipulation Classification.}
Manipulation classification commonly addresses $(i)$ misinformative claims with manipulated content; $(ii)$ correct claims accompanied by manipulated content (e.g.\ to increase credibility).
Many methods exist to manipulate text, visual and audio content. While some 
require more knowledge to use (e.g.\ speech synthesis), other manipulations can be achieved with simple tools (e.g.\ changing speed of videos)~\citep{paris2019deepfakes}.
Different terms have been used for manipulated content in recent years.
A \emph{deepfake} is commonly defined as ``the product of artificial intelligence (AI) applications that [...] create fake videos that appear authentic''~\citep{MarasA19},
with popular examples including  realistic-looking videos where the speaker's voice or face is modified 
~\citep{paris2019deepfakes}. 
On the other hand, \textit{cheap fake} defines manipulated content created through more accessible methods 
~\citep{paris2019deepfakes}, e.g.\ changing captions or speed of videos~\citep{DBLP:conf/mir/LaTTTDD22}. 
The term \emph{fauxtography} was first coined in journalism for images manipulated to ``convey a questionable (or outright false) sense of the events they seem to depict''~\citep{Cooper07, KalbC07}. 
Other terms used in the  literature for manipulated content are \emph{fake}~\citep{cheema-etal-2022-mm}, \emph{forgery}~\citep{CozzolinoRTNV21}, and \emph{splice}~\citep{ZampoglouPK15}. 
%

\paragraph{Stage $3.2$: Out-of-context Classification.}
Using unchanged content 
out-of-context is one of the most common and easiest methods to create multimodal misinformation~\citep{luo-etal-2021-newsclippings, DBLP:journals/corr/abs-2101-06278}, and involves (possibly misinformative) textual claims paired with content (e.g.\ a video) taken out of context
~\citep{ZhangSGLLZA018, Abdelnabi_2022_CVPR, Garimella2020}.
Recent work has also studied the applicability of traditional multimodal misinformation detection methods to identify out-of-context content~\citep{zhang-etal-2023}.
Other terms used for combining multimodal content in a misleading way include 
\emph{cross-modal (in-) consistency}~\citep{Muller-BudackTD20} and \emph{repurposing}~\citep{luo-etal-2021-newsclippings}.


\paragraph{Stage $3.3$: Veracity Classification.}
This task is the multimodal counterpart to 
classifying the veracity of textual claims given retrieved evidence
~\citep{thorne-vlachos-2018-automated}.
%
Veracity classification of claims embedded in audio is also commonly referred to as \textit{deception detection}~\citep{DBLP:conf/asru/KopevAKN19, DBLP:journals/ieeemm/KambojHARA21}.
While 
earlier work considered
mostly claims recorded in staged setups~\citep{newman2003lying} or from court trials~\citep{Perez-RosasAMB15}, more recently real-world political debates have become popular. 
\citep{DBLP:conf/asru/KopevAKN19, DBLP:journals/ieeemm/KambojHARA21}.
\paragraph{Stage $3.4$: Justification Production.}
Different to previous research on automated justification production
~\citep{kotonya-toni-2020-explainable},
human fact-checkers also give justifications for fact-checks involving images, audios, or videos~\citep{silverman2013verification}. 
Justifications for multimodal misinformation can be grouped in three categories: $(i)$ identifying which part of the claim input is misleading (e.g.\ specific areas in a visual claim or words in a textual one) ~\citep{KouZSW20, DBLP:conf/taai/PurwantoSLYP21, LourencoP22}; $(ii)$ providing natural language justifications following human fact-checkers~\citep{Yao2022}; $(iii)$ selecting and highlighting evidence parts used for verification~\citep{atanasova-etal-2020-generating-fact, ShangKZW22}.  
Justifications serve purposes beyond explaining veracity classification,
e.g.\ human fact-checkers also use them to discuss uncertainties and potential errors -- especially needed in fact-checking for rapidly developing events~\citep{silverman2013verification}.


    


    

\section{Datasets and Modeling Approaches}
\label{sec:modeling}


\subsection{Stage $1$: Claim Detection and Extraction}

\textbf{Input.}
Typical inputs to claim detection are unimodal, including image~\citep{Garimella2020, qu2022}, audio~\citep{DBLP:conf/misdoom/MarosAV21}, and video~\citep{ShangKZ021, qi2023}, which are collected from social media platforms such as WhatsApp and TikTok (see Table~\ref{tab:detection}).
The written or spoken claim is extracted from the input at this stage before fact-checking it.

\noindent\textbf{Output.}
Claim detection is typically framed as a classification task. Models predict if a claim is checkable or check-worthy~\citep{PrabhakarGNG21, cheema-etal-2022-mm, BarronCedenoACMEGHRSNCAN23}. The verdict for factual-verifiability 
is often binary~\citep{JinCGZL17,ShangKZ021}. For check-worthiness, \citet{PrabhakarGNG21} defines three categories of multimodal claims: statistical/numerical claims, claims about world events/places/noteworthy individuals, and other factual claims. 
\citet{cheema-etal-2022-mm} extend the binary labels for textual check-worthiness~\citep{DBLP:conf/cikm/HassanLT15} with images to be considered as well. A tweet is considered check-worthy if it is potentially harmful, breaking news, or up-to-date.

\noindent\textbf{Modeling Approaches.}
Detecting claims is a challenging task due to the vast number of posts that are published every day. Existing claim detection methods primarily rely on input content since the large volume of potentially check-worthy inputs makes it difficult to retrieve and use evidence. The early multimodal method directly concatenated visual and textual features for detection~\citep{JinCGZL17, WangMJYXJSG18}. However, simple modality fusion may not be sufficient to capture the complex relationships among multimodal information. As a result, later efforts focused on jointly learning representations across modalities. For instance, \citet{KhattarG0V19} leverage a variational auto-encoder~\citep{KingmaW13} to learn a shared representation of visual and textual content. Various attention mechanisms have also been developed to fuse multimodal representations~\citep{qian2021hierarchical, wu-etal-2021-multimodal, LiuWL23, Qi23}. Another popular approach is to use graph neural networks~\citep{KipfW17} to model the interactions among different modalities~\citep{ZhengZGWZ022, SunQLZ23}.

\begin{table*}
\centering
\scalebox{0.79}{
\begin{tabular}{cccccccc}
\hline  
 \textbf{Dataset} & \textbf{Input} & \textbf{Context} & \textbf{Output} & \textbf{\#Input} & \textbf{Lang} & \textbf{Source}\\
\hline

Weibo~\citep{JinCGZL17} & Img/Txt & Meta & 2 & 9,528 & Zh & Weibo/News \\
FauxBuster~\citep{ZhangSGLLZA018} & Img/Txt & Txt/Meta & 2  & 917 & En & Twitter/Reddit \\
Exfaux~\citep{KouZSW20} & Img/Txt & Txt & 2/4 & 263 & En & Twitter/Reddit \\
MuMIN~\citep{NielsenM22} & Img/Txt & Meta & 3 & 12,914 & Mul & Twitter\\
MMClaims~\citep{cheema-etal-2022-mm} & Img/Txt & - & 4 & 3,400 & En & Twitter \\
ContrastFaux~\citep{ZongZSW23} & Img/Txt & - & 2 & 1,841 & En & Twitter/Reddit \\
CLEF2023~\citep{BarronCedenoACMEGHRSNCAN23} & Img/Txt & - & 4 & 6,000 & Mul & Twitter \\
MR2~\citep{HuGCWY23} & Img/Txt & Txt/Img/Meta & 3 & 14,700 & Mul & Twitter/Weibo \\
\hline
IndiaWApp~\citep{Garimella2020} & Img & Meta & 2  & 2,500 & Mul & WhatsApp \\
DisinfoMeme~\citep{qu2022} & Img & - & 2  & 1,170 & En & Reddit \\
WhatsApp~\citep{DBLP:conf/misdoom/MarosAV21} & Aud & Meta & 2  & 42,689 & Pt & WhatsApp \\
TikTok~\citep{ShangKZ021} & Vid & Txt/Meta & 2 & 891 & En & TikTok \\
COVID-VTS~\citep{LiuYS23} & Vid & Txt/Aud & 2 & 10,000 & En & Twitter
\\
FakeSV~\citep{qi2023} & Vid & Txt/Meta & 2 & 3,654 & Zh & TikTok/Kuai
\\
MisDissem~\citep{ResendeMSMVAB19} & Vid/Aud/Img/Text & Meta & 2 & 121,781 & Pt & WhatsApp \\
CheckMate~\citep{PrabhakarGNG21} & Vid/Img/Text & Meta & 3 & 2,200 & Hi & Sharechat \\

\hline
\end{tabular}}
\vspace{-0.5em}
\caption{\label{tab:datasets_claim_detection} Datasets for claim detection. Img, Txt, Vid, Aud, and Meta denote image, text, video, audio, and metadata, respectively. Output indicates the number classification labels. Mul indicates that the input has multiple languages.}
\label{tab:detection}
\vspace{-1em}
\end{table*}


Multimodal content can implicitly provide claims, as seen in images and videos on social media that often have accompanying text. To extract claims from visual input, OCR systems are commonly used~\citep{Garimella2020, PrabhakarGNG21}. \citet{DBLP:journals/corr/abs-2205-12617} 
use Google Vision API to identify text in memes. Claim extraction becomes more challenging when dealing with video inputs. \citet{ShangKZ021} address this challenge by extracting captions and audio chunks after sampling video frames. These captions and audio chunks were then encoded into representations to guide the visual feature extraction process. For audio inputs, \citet{DBLP:conf/misdoom/MarosAV21} use Google’s Speech-to-Text API to produce transcripts.

\subsection{Stage $2$: Evidence Retrieval}

Previous work uses different types of evidence and retrieval methods given the modalities involved. 
Evidence data and retrieval approaches can be grouped into $(i)$ content-based and $(ii)$ retrieval-based (see column \emph{evidence} in Table~\ref{tab:verdict}).  

\noindent\textbf{Content-based.}
Content-based approaches use the claim and its context (i.e.\ the same information that is used for claim detection and extraction) 
as evidence instead of retrieving additional data.
This is particularly common for audio and video misinformation (Table~\ref{tab:verdict}).
Acoustic or visual features extracted from the input are used as evidence for verdict prediction~\citep{WuKEYHSS15,YiBTMTWWF21,ISMAELALSANJARY2016565,JiangLW0L20}.
Most approaches 
use audio (or video) features and accompanying data (e.g.\ metadata, transcripts if available) as evidence to identify inconsistencies~\citep{DBLP:conf/asru/KopevAKN19, Andreas18, LiYSQL20}. 
Several datasets with image/text claims~\citep{TanPS20,luo-etal-2021-newsclippings,DBLP:journals/corr/abs-2101-06278} also do not retrieve additional evidence (Table~\ref{tab:verdict}) but rely on the given claim input or use accompanying metadata~\citep{JaiswalSAN17, SabirA0N18}. 
Metadata is also often used as evidence for verdict prediction with images as input (Table~\ref{tab:verdict}). \citet{JaiswalSAN17} and \citet{SabirA0N18} use metadata (e.g.\ image timestamps) to provide additional information. Similarly, \citet{HuhLOE18} incorporate EXIF metadata (e.g.\ camera version, focal length, resolution settings) to detect manipulation.
Image captions are also used as evidence sometimes~\citep{Shao_2023_CVPR}.

\noindent\textbf{Retrieval-based.}
Retrieved evidence external to the claim is mostly used for fact-checking text claims, text/image and image claims while audio and video fact-checks often don't retrieve additional evidence data (Table~\ref{tab:verdict}) but rely on the content of the video/audio input.
\citet{fung-etal-2021-infosurgeon} leverage a knowledge base for additional background knowledge. They first construct a knowledge graph of the input news article using its text and images. They extract entities/relations from this knowledge graph with an Information Extraction system~\citep{li-etal-2020-cross, lin-etal-2020-joint} and map the entities to Freebase~\citep{BollackerEPST08} as their background knowledge base. 
 %
Two recent datasets scrape claims from fact-checking websites, and include text/image/video from those articles as evidence~\citep{DBLP:conf/icwsm/SinghalSK22, Yao2022}.
\citet{akhtar-etal-2023-reading} used chart images as evidence to verify textual claims. 
To determine if an image is used out-of-context, previous works also use \emph{(reverse) image search}~\citep{Muller-BudackTD20, Abdelnabi_2022_CVPR}, to find evidence sources with images similar to or same as the claim image. 
\citet{Muller-BudackTD20} query 
search engines and the \emph{WikiData} knowledge graph
using named entities from the claim text.
\citet{Abdelnabi_2022_CVPR} use the claim image caption and the image itself as query.

\begin{table*}
\centering
\scalebox{0.66}{
\begin{tabular}{cccccccc}
\hline  
 \textbf{Dataset} & \textbf{Input} & \textbf{Evidence} & \textbf{Output} & \textbf{Tasks} & \textbf{\#Input} & \textbf{Lang} & \textbf{Source} \\
\hline
MAIM~\citep{JaiswalSAN17} & Img/Txt & Meta & 2 & O & 239,968 & En & Flickr \\
MEIR~\citep{SabirA0N18} & Img/Txt & Meta & 2 & O & 140,096 & En & Flickr \\
TNews~\citep{Muller-BudackTD20} & Img/Txt & Img & 2 & O  & 72,561 & En & News \\
News400~\citep{Muller-BudackTD20} & Img/Txt & Img &  2 & O  & 400 & En/De & News \\
NeuralNews~\citep{TanPS20} & Img/Txt & - & 4 & O  & 128,000 & En & Grover/GoodNews \\
COSMOS~\citep{DBLP:journals/corr/abs-2101-06278} & Img/Txt & - & 2 & O & 201,700 & En & News/Snopes \\
NewsCLIPings~\citep{luo-etal-2021-newsclippings} & Img/Txt & - & 2 & O & 988,283 & En & CLIP/VisualNews \\
InfoSurgeon~\citep{fung-etal-2021-infosurgeon} & Img/Txt & KB/Meta & 2 & O & 30,000 & En & VoA \\
Factify~\citep{factify2} & Img/Txt & Txt & 5 & O & 50,000 & En & Twitter \\
\hline
FakingSandy~\citep{Gupta13} & Img & Txt/Meta & 2 & M & 16,117 & - & Twitter \\
MediaEval~\citep{boididou2014challenges} & Img & Txt/Meta & 2 & M & 13,924 & - & Twitter \\
In-the-Wild~\citep{HuhLOE18} & Img & Meta & 2 & M & 201 & - & Reddit/Onion\\
PS-Battles~\citep{Heller18} & Img & Txt/Meta &  2 & M  & 103,028 & - & Reddit  \\
DGM~\citep{Shao_2023_CVPR} & Img & Txt &  2 & M  & 230,000 & - & News  \\
VTD~\citep{ISMAELALSANJARY2016565} & Vid & - & 2 & M & 33 & En & YouTube  \\
Faceforensics~\citep{Andreas18} & Vid & - & 2 & M & 1,004 & En & YouTube \\
DeepfakeDetect~\citep{GueraD18} & Vid & - & 2 & M & 600 & En & Vid Webs./HOHA \\
DFDC~\citep{Brian2019} & Vid & - & 2 & M & 128,154 & En & Recorded \\
DeeperForensics-1.0~\citep{JiangLW0L20} & Vid & - & 2 & M & 60,000 & En & Recorded \\
Celeb-DF~\citep{LiYSQL20} & Vid & - & 2 & M & 6,229 & En & YouTube \\
KoDF~\citep{DBLP:conf/iccv/KwonYNPC21} & Vid & - & 2 & M & 237,942 & Ko & Recorded \\
DF-Platter~\citep{Narayan_2023_CVPR} & Vid & - & 2 & M & 133,260 & En & YouTube \\
ASVspoof~\citep{WuKEYHSS15} & Aud & - & 2 & M & 16,375 & En & SAS \\
Phonespoof~\citep{LavrentyevaNVMM19} & Aud & - & 2 & M & 34,407 & En & ASVspoof \\
FoR~\citep{ReimaoT19} & Aud & - & 2 & M & 53,868 & En & TTS Systems \\
DeepSonar~\citep{WangJHGXML20} & Aud & - & 2 & M & 18,614 & En/Zh & TTS Systems/VCC \\
HAD~\citep{YiBTMTWWF21} & Aud & - & 3 & M & 88,035 & Zh & AISHELL-3 \\
FakeAVCeleb~\citep{DBLP:conf/nips/KhalidTKW21} & Vid/Aud & - & 4 & M & 20,000 & En & VoxCeleb2 \\

\hline

MedVideo~\citep{DBLP:conf/icmi/HouPLM19} & Vid & - & 2 & VC & 250 & En & YouTube \\
CLEF2018 Audio~\citep{DBLP:conf/asru/KopevAKN19} & Aud & Meta & 3 & VC & 286 & En & Debates \\
FactDrill~\citep{DBLP:conf/icwsm/SinghalSK22} & Txt & Vid/Aud/Img/Txt/Meta & 5 & VC  & 22,435 & Mul & FC webs. \\
MMM~\citep{GuptaKAGE22} & Txt & Img/Meta & 2 & VC  &  10,473 & Mul & FC webs. \\
ChartFC~\citep{akhtar-etal-2023-reading} & Txt & Img & 2 & VC  & 15,886 & En & TabFact \\
Fauxtography~\citep{zlatkova-etal-2019-fact} & Img/Txt & Meta & 2 & VC & 1,233 & En & Snopes/Reuters \\
MOCHEG~\citep{Yao2022} & Img/Txt & Img/Txt & 3 & VC & 21,184 & En & FC webs. \\

r/Fakeddit~\citep{NakamuraLW20} & Img/Txt & Meta & 2/3/6 & O/M/VC & 1,063,106 & En & Reddit \\

\hline
\end{tabular}}
\vspace{-0.5em}
\caption{\label{tab:manipulation_fabric_ooc} Datasets for manipulation, out-of-context, and veracity classification. O, M and VC denote out-of-context, manipulation and veracity classification, respectively. Mul indicates the input has multiple languages. }
\label{tab:verdict}
\vspace{-1em}
\end{table*}

\subsection{Stage $3$: Verdict Prediction}

As introduced in Section~\ref{sec:taskformulation}, the verdict prediction stage includes manipulation, out-of-context, and veracity classification as sub-tasks.

\noindent\textbf{Input.}
As shown in Table~\ref{tab:verdict}, inputs of \textbf{manipulation classification} datasets usually focus on one modality. 
For dataset creation, manipulated images are often collected from social media platforms such as Twitter, Reddit, and YouTube~\citep{Gupta13,Heller18}.
For verdict prediction datasets with videos, in addition to social media~\citep{ISMAELALSANJARY2016565}, film clips~\citep{GueraD18}, facial expressions~\citep{Andreas18}, and interviews~\citep{LiYSQL20} are used. Some works record videos to simulate real-world scenarios~\citep{Brian2019,JiangLW0L20,DBLP:conf/iccv/KwonYNPC21}.
To create datasets of manipulated content,  altering methods
based on GANs have also been applied in earlier works \cite{ZakharovSBL19,NirkinKH19,KarrasLA19}.
For audio manipulations, most benchmarks~\citep{WuKEYHSS15, KinnunenSDTEYL17,ReimaoT19,WangJHGXML20,YiBTMTWWF21} use speech synthesis and voice conversion algorithms to collect manipulated audios. To assess real-world audio  manipulations, \citet{LavrentyevaNVMM19} emulate realistic telephone channels.

Most \textbf{out-of-context classification} datasets have image-caption pairs as input (Table~\ref{tab:verdict}). \citet{JaiswalSAN17} replace captions of Flickr images to get mismatched pairs. As replacing the 
entire caption can be easy to detect, later efforts~\citep{SabirA0N18,Muller-BudackTD20} propose to change specific entities in them.  \citet{luo-etal-2021-newsclippings} show that such text manipulations introduce linguistic biases and can be solved without the images. They use CLIP~\citep{DBLP:conf/icml/RadfordKHRGASAM21} to filter out pairs that do not require multimodal modeling. 
Popular sources for out of context datasets with text and image claims include Flickr and news/fact-checking websites~\citep{DBLP:journals/corr/abs-2101-06278, JaiswalSAN17, SabirA0N18}.

The primary input to multimodal \textbf{veracity classification} is the content-to-be-checked itself -- typically text, audio or video in past benchmarks.
\citet{DBLP:conf/asru/KopevAKN19} include verified speeches from the CLEF-2018 Task 2~\citep{NakovBESMZAKM18} while \citet{DBLP:conf/icmi/HouPLM19} collect videos about prostate cancer verified by urologists. \citet{zlatkova-etal-2019-fact} and \citet{Yao2022} collect viral images with texts verified by dedicated agencies.  \citet{NakamuraLW20} collect image-text pairs from Reddit via distant supervision, e.g.\ labeling a post from the subreddit ``fakefacts'' as \textit{misleading} and 
from ``photoshopbattles'' as \textit{manipulated}. 
For veracity classification of spoken claims, real-world political debates are popular sources for claims~\citep{DBLP:conf/asru/KopevAKN19, DBLP:journals/ieeemm/KambojHARA21}. 
For example, \citet{DBLP:conf/asru/KopevAKN19} and \citet{DBLP:journals/ieeemm/KambojHARA21} use claims labelled by fact checking organizations,  and
video recordings as well as transcripts of the respective political debates. 

\noindent\textbf{Output.} Most manipulation and out-of-context classification datasets use binary labels: ``out-of-context/not out-of-context''~\citep{Muller-BudackTD20, luo-etal-2021-newsclippings}, ``pristine/falsified''~\citep{boididou2014challenges, Heller18}, ``manipulation/no manipulation'' \citep{Brian2019,LiYSQL20}. Following fact-checkers, veracity classification datasets~\citep{DBLP:conf/icwsm/SinghalSK22,NakamuraLW20} sometimes employ multi-class labels to represent degrees of truthfulness (e.g.\ true, mostly-true, half-true) (see Table~\ref{tab:verdict}). 
\citet{mishra2022factify} adopt labels to denote the entailment between different claim and evidence modalities, e.g.\ the label \textit{support text} denotes that only the textual part of the evidence supports the claim but not the accompanying image while \textit{support multimodal} includes both modalities.

\noindent\textbf{Modeling Approaches.}
To detect visual manipulations, 
early approaches mostly use CNN models, such as VGG16~\citep{AmeriniGCB19, DBLP:conf/cvpr/DangLS0020}, ResNet~\citep{AmeriniGCB19, DBLP:conf/cvpr/SabirCJAMN19}, and InceptionV3~\citep{GueraD18}. 
Some works extend them to capture temporal aspects of video \textbf{manipulation classification}.
\citet{AmeriniGCB19} adopt optical flow fields to capture the correlation between consequent video frames and detect dissimilarities caused by manipulation. 
\citet{GueraD18} 
model temporal information with an LSTM model and a sequence of features vectors per video frame to classify manipulated videos. 
\citet{DBLP:conf/cvpr/SabirCJAMN19} similarly extract features for video frames and detect discrepancies between frames using a recurrent convolution network. 
Some recent models also integrate transformer-based components 
~\citep{VaswaniSPUJGKP17, DBLP:conf/iccv/ZhengB0ZW21}.
For example, \citet{DBLP:conf/mir/WangWOHCJL22} combine CNNs and vision transformers (ViTs)~\citep{DosovitskiyB0WZ21} while \citet{DBLP:journals/corr/abs-2102-11126} introduce a multi-scale ViT with variable patch sizes. 

Models for \textbf{out-of-context} and \textbf{veracity classification} typically consist of unimodal encoders, a fusion component to obtain joint, multimodal representations, and a classification component.
To obtain text representations, early approaches used combinations of word2vec models~\citep{MikolovCCD13}, LSTMs~\citep{HochreiterS97}, and TF-IDF scores for n-grams~\citep{JinCGZL17, 9182398, DBLP:conf/icmi/HouPLM19}. More recent efforts use pretrained language models
~\citep{fung-etal-2021-infosurgeon, DBLP:journals/corr/abs-2101-06278, DBLP:conf/mir/LaTTTDD22}. 
To encode visual data, many approaches first detect objects in visual content using a Mask R-CNN model~\citep{HeGDG17} before extracting visual features~\citep{DBLP:journals/corr/abs-2101-06278, DBLP:conf/mir/LaTTTDD22, ShangKZW22}.
Visual representations for images and videos are commonly obtained using CNN models such as ResNet~\citep{HeZRS16,Garimella2020,Abdelnabi_2022_CVPR}, VGG~\citep{SimonyanZ14a, JinCGZL17, SabirA0N18}, 
and Inception~\citep{SzegedyWYSRADVR15, GueraD18, DBLP:conf/ijcnn/RoyE21}. 
To obtain audio features for voice quality, loudness, and tonality, \citet{ShangKZ021} extract the Mel-frequency cepstral coefficient, \citet{DBLP:conf/asru/KopevAKN19} use the INTERSPEECH 2013 ComParE feature set~\citep{DBLP:conf/mm/EybenWGS13}, and \citet{DBLP:conf/icmi/HouPLM19} use the openEAR toolkit~\citep{DBLP:conf/acii/EybenWS09}.
Various approaches have been used to obtain \textbf{multimodal representations}. Early fusion, which joins representations immediately after the encoding step~\citep{DBLP:journals/pami/BaltrusaitisAM19} is more common~\citep{DBLP:journals/corr/abs-2101-06278, 9182398, DBLP:conf/mir/LaTTTDD22} than late fusion~\citep{Yao2022}. Moreover, model-agnostic methods (e.g.\ concatenation and dot product) are more prevalent~\citep{DBLP:journals/corr/abs-2101-06278, DBLP:conf/asru/KopevAKN19, JinCGZL17, DBLP:conf/mir/LaTTTDD22} than model-based approaches (e.g.\ neural networks)~\citep{JaiswalSAN17, ShangKZW22}. 
Also popular for out-of-context classification are \emph{cross-modality checks} that compare modalities present in a claim to each other, e.g.\ a video and its caption \citep{Muller-BudackTD20, fung-etal-2021-infosurgeon}.

\subsection{Stage $3$: Justification Production}

A small number of datasets is available for multimodal justification production.
Previous work can be grouped into two categories: $(1)$ highlighting parts of the input, and $(2)$ generating natural language justifications.

\noindent\textbf{Highlighting Input.}
The first category highlights input parts as justification which contribute to models' results. 
A popular approach for this are Graph Neural Networks~\citep{KipfW17}. 
Several papers encode multimodal data as graph elements, combining entities and their relations in and between modalities. Models are trained to detect inconsistencies between 
different modalities, or to detect relations (i.e., between entities) that may be misinformative. This detection could be based on the local graph structure, or on an external knowledge base~\citep{fung-etal-2021-infosurgeon, ShangKZW22, KouZSW20}. Highlighted entities and relations serve as explanations for the potential misinformativeness of the entire graph. 
Conversely, \citet{Zhou_2018_CVPR} and \citet{DBLP:conf/cvpr/0001AN19} use a multitask model for manipulation classification and identification of manipulated regions. Rather than labeled data, some papers rely on attention mechanisms to highlight areas as explanations. \citet{DBLP:conf/icpr/BonettiniCMBBT20, DBLP:conf/cvpr/DangLS0020} use this approach to highlight manipulated image regions; \citet{DBLP:conf/taai/PurwantoSLYP21} also include captions. 

\noindent\textbf{Natural Language Justifications.}
\citet{Yao2022} recently introduced a multimodal dataset with natural language justifications.
They scrape text and visual content from web pages referenced by fact-checking articles. 
The dataset includes summaries in the fact-checking articles as gold justifications for the verdicts. However, such a setting is not realistic, as fact-checking articles are not available when verifying a new claim. 
\section{Challenges and Future Directions}
\label{sec:challenges}

\paragraph{Claim extraction from multimodal content.} 
Multimodal claims, e.g.\ manipulated videos, are often embedded in specific contexts and framed as (part of) larger stories. For example, countering the misinformation
in Figure~\ref{fig:deepfake} requires not only classifying if the image is manipulated, but understanding that it depicts the arrest of the former president in one of the cases he is being charged in. Only then can relevant evidence data be extracted and used to verify the story of Trump's arrest.
To determine what is being claimed is a challenging first step in multimodal automated fact-checking. 
However, current efforts for multimodal claim extraction are limited to text extraction from visual content or transcribing audios and videos~\citep{DBLP:journals/corr/abs-2205-12617, Garimella2020, DBLP:conf/misdoom/MarosAV21}. 
Addressing this challenge will require modeling approaches to effectively align and integrate all modalities present in and around the claim.
For example, methods for pixel-based language modeling have recently been introduced to better align visually situated language with image content~\citep{LeeJTHLEKSCT2022}. Such approaches considering modalities beyond text and vision for multimodal data alignment can be useful for claim extracting from multimodal input. 


 
\paragraph{Multimodal evidence retrieval.} Evidence retrieval for audio and video fact-checking remains a major challenge. Different to other modalities, they cannot be easily searched on the web or social media networks~\citep{silverman2013verification}. 
Fact-checkers often use text 
accompanying the videos to find evidence~\citep{silverman2013verification}. Reverse image search engines, e.g.\ Google Lens or TinEye, require screenshots from the video as input -- and thus require the correct timeframe, which can be challenging to extract.
A dedicated adversary can render current tools very difficult to use.
Very often evidence for image or audio fact-checking is retrieved using text accompanying them , e.g.\ metadata, social media comments, or captions~\citep{Gupta13, HuhLOE18, Muller-BudackTD20, DBLP:conf/asru/KopevAKN19}. While incorporating the textual information and the other modality (e.g. audio/image) in retrieval would provide more information, this is missing currently. 
How to best retrieve evidence data that is non-textual or has a different modality than the claim, also remains a challenge. 

\paragraph{
Multilinguality and multimodality.} While there is increasing work on multilingual fact-checking~\citep{gupta-srikumar-2021-x, Shahi2020FakeCovidA, HammouchiG22}, it is mostly limited to text-only benchmarks and models. Surveying benchmarks for different pipeline stages (Figure~\ref{fig:overviewtask}), we found limited multimodal datasets for non-English languages (see Table~\ref{tab:manipulation_fabric_ooc}). Previous work on multilingual multimodality shows that training and testing on English data alone introduces biases, 
as models fail to capture concepts and images prevalent in other languages and cultures~\citep{liu-etal-2021-visually}. 
Moreover, some types of multimodal misinformation exploit cross-lingual sources to mislead, e.g.\ images or videos  from non-English newspapers appearing as out-of-context data for English multimodal misinformation~\citep{silverman2013verification}.
To prevent false conclusions and biases, 
it is thus necessary to take approaches that are both multimodal \textit{and} multilingual
~\citep{ruder-etal-2022-square}. Construction of large-scale multimodal, multilingual AFC datasets would facilitate futures research in this direction, similar to benchmarks and shared tasks created for automated fact-checking tasks in English~\citep{thorne-etal-2018-fact, Facitfy23}.

\paragraph{Generalizing detection of visual manipulations.}
The recent popularity of diffusion models (DMs) for visual manipulation have raised questions regarding the generalizability of manipulation detectors developed for earlier models (e.g. GANs~\citep{GoodfellowPMXWO20}). 
Detection models are biased towards specific manipulation models and struggle to generalize~\citep{wu-etal-generalizable, ricker-etal-2022-towards}.
A recent study~\citep{ricker-etal-2022-towards} shows that detectors initially developed for GANs, have average performance drops of around $15\%$ for image by DMs.
While new detection approaches for DM manipulations are already being developed~\citep{GuarneraGB2023, WuLO2023}, the question how to generalize and increase robustness of manipulation detectors for potential future manipulation models remains open. 
Potential solutions can include evidence-based approaches, 
where the manipulated content is used to retrieve evidence data (e.g. the original video or counterfactual evidence) to prove the manipulation.

\paragraph{Justifications for multimodal fact-checking.} While explainable fact-checking has received attention recently~\citep{kotonya-toni-2020-explainable-automated, atanasova-etal-2020-generating-fact}, there is limited work on producing justifications for multimodal content. 
Previous efforts on multimodal justification production have mostly focused on highlighting parts of the input to increase \emph{interpretability}~\citep{KouZSW20, ShangKZW22}. 
Natural language justifications that explain the fact-check of multimodal claims so that it is accessible to non-technical have not been developed yet. 
To develop solutions, we first need appropriate benchmarks to measure progress. 
Moreover, with the recent advances of neural models for visual and audio generation and editing, another so far unexplored direction presents itself: editing input images/videos/audios or generating entirely content to explain fact-checking results. This could include, for example, the generation of infographics or video clips to explanation fact-checks. Such a system, especially if guided by human fact-checkers~\citep{NakovCHAEBPSM21}, would be a potent tool. 
As noted in  \citet{lewandowsky_debunking_2020}, ``well-designed graphs, videos, photos, and other semantic aids can be helpful to convey corrections involving complex or statistical information clearly and concisely''.
%


\section{Conclusion}

We survey research on multimodal automated fact-checking and introduce a framework that combines and organizes tasks introduced in various communities studying misinformation. We discuss common terms and definitions in context of our framework. We further study popular benchmarks and modeling approaches, and discuss promising directions for future research. 

\section*{Limitations}

While we cite many datasets and modeling approaches for multimodal fact-checking, we describe most of them only briefly due to space constraints. Our aim was to provide an overview of multimodal fact-checking and organise previous works in a framework. 
Moreover, the presented survey focuses primarily on four modalities. While there are other modalities we could have included, we concentrated on those prevalent in real-world fact-checking that have not been discussed as part of a fact-checking framework in previous surveys.

\section*{Ethics Statement}

As we mention in Section~\ref{sec:challenges}, most datasets for multimodal fact-checking tasks are available only in English. Thus, models are evaluated based on their performance on English benchmarks only. 
This can lead to a distorted view about advancements on multimodal automated fact-checking as it is limited to a single language out of more than $7000$ world languages. While we call for future work on a variety of languages, this survey provides an overview on the state-of-the-art of mostly-English research efforts. 
Finally, we want to point out that multimodal fact-checking works we cite in this survey might include misleading statements or images given as examples.

\section*{Acknowledgements}
Zhijiang Guo, Michael Schlichtkrull and Andreas Vlachos are supported by the ERC grant AVeriTeC (GA 865958).
This paper is produced as part of the MuseIT project which has been co-funded by the EU under the Grant Agreement number 101061441. MuseIT has supported the work of Mubashara Akhtar. Views and opinions expressed are however those of the author(s) only and do not necessarily reflect those of the European Union or the European Research Executive Agency, REA. Neither the EU nor the granting authority can be held responsible for them.

\bibliography{anthology,custom}

\begin{thebibliography}{157}
\expandafter\ifx\csname natexlab\endcsname\relax\def\natexlab#1{#1}\fi

\bibitem[{Abdali(2022)}]{DBLP:journals/corr/abs-2203-13883}
Sara Abdali. 2022.
\newblock \href {https://doi.org/10.48550/arXiv.2203.13883} {Multi-modal
  misinformation detection: Approaches, challenges and opportunities}.
\newblock \emph{CoRR}, abs/2203.13883.

\bibitem[{Abdelnabi et~al.(2022)Abdelnabi, Hasan, and
  Fritz}]{Abdelnabi_2022_CVPR}
Sahar Abdelnabi, Rakibul Hasan, and Mario Fritz. 2022.
\newblock Open-domain, content-based, multi-modal fact-checking of
  out-of-context images via online resources.
\newblock In \emph{Proceedings of the IEEE/CVF Conference on Computer Vision
  and Pattern Recognition (CVPR)}, pages 14940--14949.

\bibitem[{Agarwal et~al.(2019)Agarwal, Farid, Gu, He, Nagano, and
  Li}]{DBLP:conf/cvpr/AgarwalFGHN019}
Shruti Agarwal, Hany Farid, Yuming Gu, Mingming He, Koki Nagano, and Hao Li.
  2019.
\newblock \href
  {http://openaccess.thecvf.com/content\_CVPRW\_2019/html/Media\_Forensics/Agarwal\_Protecting\_World\_Leaders\_Against\_Deep\_Fakes\_CVPRW\_2019\_paper.html}
  {Protecting world leaders against deep fakes}.
\newblock In \emph{{IEEE} Conference on Computer Vision and Pattern Recognition
  Workshops, {CVPR} Workshops 2019, Long Beach, CA, USA, June 16-20, 2019},
  pages 38--45. Computer Vision Foundation / {IEEE}.

\bibitem[{Akhtar et~al.(2022)Akhtar, Cocarascu, and
  Simperl}]{akhtar-etal-2022-pubhealthtab}
Mubashara Akhtar, Oana Cocarascu, and Elena Simperl. 2022.
\newblock \href {https://doi.org/10.18653/v1/2022.findings-naacl.1}
  {{P}ub{H}ealth{T}ab: {A} public health table-based dataset for evidence-based
  fact checking}.
\newblock In \emph{Findings of the Association for Computational Linguistics:
  NAACL 2022}, pages 1--16, Seattle, United States. Association for
  Computational Linguistics.

\bibitem[{Akhtar et~al.(2023)Akhtar, Cocarascu, and
  Simperl}]{akhtar-etal-2023-reading}
Mubashara Akhtar, Oana Cocarascu, and Elena Simperl. 2023.
\newblock \href {https://aclanthology.org/2023.findings-eacl.30} {Reading and
  reasoning over chart images for evidence-based automated fact-checking}.
\newblock In \emph{Findings of the Association for Computational Linguistics:
  EACL 2023}, pages 399--414, Dubrovnik, Croatia. Association for Computational
  Linguistics.

\bibitem[{Alam et~al.(2022)Alam, Cresci, Chakraborty, Silvestri, Dimitrov,
  Martino, Shaar, Firooz, and Nakov}]{alam-etal-2022-survey}
Firoj Alam, Stefano Cresci, Tanmoy Chakraborty, Fabrizio Silvestri, Dimiter
  Dimitrov, Giovanni Da~San Martino, Shaden Shaar, Hamed Firooz, and Preslav
  Nakov. 2022.
\newblock \href {https://aclanthology.org/2022.coling-1.576} {A survey on
  multimodal disinformation detection}.
\newblock In \emph{Proceedings of the 29th International Conference on
  Computational Linguistics}, pages 6625--6643, Gyeongju, Republic of Korea.
  International Committee on Computational Linguistics.

\bibitem[{Aly et~al.(2021)Aly, Guo, Schlichtkrull, Thorne, Vlachos,
  Christodoulopoulos, Cocarascu, and Mittal}]{DBLP:conf/nips/AlyGST00CM21}
Rami Aly, Zhijiang Guo, Michael~Sejr Schlichtkrull, James Thorne, Andreas
  Vlachos, Christos Christodoulopoulos, Oana Cocarascu, and Arpit Mittal. 2021.
\newblock \href
  {https://datasets-benchmarks-proceedings.neurips.cc/paper/2021/hash/68d30a9594728bc39aa24be94b319d21-Abstract-round1.html}
  {{FEVEROUS:} fact extraction and verification over unstructured and
  structured information}.
\newblock In \emph{Proceedings of the Neural Information Processing Systems
  Track on Datasets and Benchmarks 1, NeurIPS Datasets and Benchmarks 2021,
  December 2021, virtual}.

\bibitem[{Amerini et~al.(2019)Amerini, Galteri, Caldelli, and
  Bimbo}]{AmeriniGCB19}
Irene Amerini, Leonardo Galteri, Roberto Caldelli, and Alberto~Del Bimbo. 2019.
\newblock \href {https://doi.org/10.1109/ICCVW.2019.00152} {Deepfake video
  detection through optical flow based {CNN}}.
\newblock In \emph{2019 {IEEE/CVF} International Conference on Computer Vision
  Workshops, {ICCV} Workshops 2019, Seoul, Korea (South), October 27-28, 2019},
  pages 1205--1207. {IEEE}.

\bibitem[{Amri et~al.(2021)Amri, Sallami, and
  A{\"{\i}}meur}]{DBLP:conf/fps/AmriSA21}
Sabrine Amri, Dorsaf Sallami, and Esma A{\"{\i}}meur. 2021.
\newblock \href {https://doi.org/10.1007/978-3-031-08147-7\_12} {{EXMULF:} an
  explainable multimodal content-based fake news detection system}.
\newblock In \emph{Foundations and Practice of Security - 14th International
  Symposium, {FPS} 2021, Paris, France, December 7-10, 2021, Revised Selected
  Papers}, volume 13291 of \emph{Lecture Notes in Computer Science}, pages
  177--187. Springer.

\bibitem[{Aneja et~al.(2021)Aneja, Bregler, and
  Nie{\ss}ner}]{DBLP:journals/corr/abs-2101-06278}
Shivangi Aneja, Christoph Bregler, and Matthias Nie{\ss}ner. 2021.
\newblock \href {http://arxiv.org/abs/2101.06278} {Catching out-of-context
  misinformation with self-supervised learning}.
\newblock \emph{CoRR}, abs/2101.06278.

\bibitem[{Atanasova et~al.(2020)Atanasova, Simonsen, Lioma, and
  Augenstein}]{atanasova-etal-2020-generating-fact}
Pepa Atanasova, Jakob~Grue Simonsen, Christina Lioma, and Isabelle Augenstein.
  2020.
\newblock \href {https://doi.org/10.18653/v1/2020.acl-main.656} {Generating
  fact checking explanations}.
\newblock In \emph{Proceedings of the 58th Annual Meeting of the Association
  for Computational Linguistics}, pages 7352--7364, Online. Association for
  Computational Linguistics.

\bibitem[{Baltrusaitis et~al.(2019)Baltrusaitis, Ahuja, and
  Morency}]{DBLP:journals/pami/BaltrusaitisAM19}
Tadas Baltrusaitis, Chaitanya Ahuja, and Louis{-}Philippe Morency. 2019.
\newblock \href {https://doi.org/10.1109/TPAMI.2018.2798607} {Multimodal
  machine learning: {A} survey and taxonomy}.
\newblock \emph{{IEEE} Trans. Pattern Anal. Mach. Intell.}, 41(2):423--443.

\bibitem[{Barr{\'{o}}n{-}Cede{\~{n}}o et~al.(2023)Barr{\'{o}}n{-}Cede{\~{n}}o,
  Alam, Caselli, Martino, Elsayed, Galassi, Haouari, Ruggeri, Stru{\ss}, Nandi,
  Cheema, Azizov, and Nakov}]{BarronCedenoACMEGHRSNCAN23}
Alberto Barr{\'{o}}n{-}Cede{\~{n}}o, Firoj Alam, Tommaso Caselli, Giovanni
  Da~San Martino, Tamer Elsayed, Andrea Galassi, Fatima Haouari, Federico
  Ruggeri, Julia~Maria Stru{\ss}, Rabindra~Nath Nandi, Gullal~S. Cheema,
  Dilshod Azizov, and Preslav Nakov. 2023.
\newblock \href {https://doi.org/10.1007/978-3-031-28241-6\_59} {The
  {CLEF-2023} checkthat! lab: Checkworthiness, subjectivity, political bias,
  factuality, and authority}.
\newblock In \emph{Advances in Information Retrieval - 45th European Conference
  on Information Retrieval, {ECIR} 2023, Dublin, Ireland, April 2-6, 2023,
  Proceedings, Part {III}}, volume 13982 of \emph{Lecture Notes in Computer
  Science}, pages 506--517. Springer.

\bibitem[{Biamby et~al.(2022)Biamby, Luo, Darrell, and
  Rohrbach}]{biamby-etal-2022-twitter}
Giscard Biamby, Grace Luo, Trevor Darrell, and Anna Rohrbach. 2022.
\newblock \href {https://doi.org/10.18653/v1/2022.naacl-main.110}
  {{T}witter-{COMM}s: Detecting climate, {COVID}, and military multimodal
  misinformation}.
\newblock In \emph{Proceedings of the 2022 Conference of the North American
  Chapter of the Association for Computational Linguistics: Human Language
  Technologies}, pages 1530--1549, Seattle, United States. Association for
  Computational Linguistics.

\bibitem[{Boididou et~al.(2014)Boididou, Papadopoulos, Kompatsiaris,
  Schifferes, and Newman}]{boididou2014challenges}
Christina Boididou, Symeon Papadopoulos, Yiannis Kompatsiaris, Steve
  Schifferes, and Nic Newman. 2014.
\newblock Challenges of computational verification in social multimedia.
\newblock In \emph{Proceedings of the 23rd International Conference on World
  Wide Web}, pages 743--748.

\bibitem[{Bollacker et~al.(2008)Bollacker, Evans, Paritosh, Sturge, and
  Taylor}]{BollackerEPST08}
Kurt~D. Bollacker, Colin Evans, Praveen~K. Paritosh, Tim Sturge, and Jamie
  Taylor. 2008.
\newblock \href {https://doi.org/10.1145/1376616.1376746} {Freebase: a
  collaboratively created graph database for structuring human knowledge}.
\newblock In \emph{Proceedings of the {ACM} {SIGMOD} International Conference
  on Management of Data, {SIGMOD} 2008, Vancouver, BC, Canada, June 10-12,
  2008}, pages 1247--1250. {ACM}.

\bibitem[{Bonettini et~al.(2020)Bonettini, Cannas, Mandelli, Bondi, Bestagini,
  and Tubaro}]{DBLP:conf/icpr/BonettiniCMBBT20}
Nicol{\`{o}} Bonettini, Edoardo~Daniele Cannas, Sara Mandelli, Luca Bondi,
  Paolo Bestagini, and Stefano Tubaro. 2020.
\newblock \href {https://doi.org/10.1109/ICPR48806.2021.9412711} {Video face
  manipulation detection through ensemble of cnns}.
\newblock In \emph{25th International Conference on Pattern Recognition, {ICPR}
  2020, Virtual Event / Milan, Italy, January 10-15, 2021}, pages 5012--5019.
  {IEEE}.

\bibitem[{Cao et~al.(2020)Cao, Qi, Sheng, Yang, Guo, and
  Li}]{DBLP:journals/corr/abs-2003-05096}
Juan Cao, Peng Qi, Qiang Sheng, Tianyun Yang, Junbo Guo, and Jintao Li. 2020.
\newblock \href {http://arxiv.org/abs/2003.05096} {Exploring the role of visual
  content in fake news detection}.
\newblock \emph{CoRR}, abs/2003.05096.

\bibitem[{Cheema et~al.(2022)Cheema, Hakimov, Sittar, M{\"u}ller-Budack, Otto,
  and Ewerth}]{cheema-etal-2022-mm}
Gullal~Singh Cheema, Sherzod Hakimov, Abdul Sittar, Eric M{\"u}ller-Budack,
  Christian Otto, and Ralph Ewerth. 2022.
\newblock \href {https://doi.org/10.18653/v1/2022.findings-naacl.72}
  {{MM}-claims: A dataset for multimodal claim detection in social media}.
\newblock In \emph{Findings of the Association for Computational Linguistics:
  NAACL 2022}, pages 962--979, Seattle, United States. Association for
  Computational Linguistics.

\bibitem[{Chen et~al.(2020)Chen, Wang, Chen, Zhang, Wang, Li, Zhou, and
  Wang}]{DBLP:conf/iclr/ChenWCZWLZW20}
Wenhu Chen, Hongmin Wang, Jianshu Chen, Yunkai Zhang, Hong Wang, Shiyang Li,
  Xiyou Zhou, and William~Yang Wang. 2020.
\newblock \href {https://openreview.net/forum?id=rkeJRhNYDH} {Tabfact: {A}
  large-scale dataset for table-based fact verification}.
\newblock In \emph{8th International Conference on Learning Representations,
  {ICLR} 2020, Addis Ababa, Ethiopia, April 26-30, 2020}. OpenReview.net.

\bibitem[{Cooper(2007)}]{Cooper07}
Stephen Cooper. 2007.
\newblock A concise history of the fauxtography blogstorm in the 2006 lebanon
  war.
\newblock \emph{American Communication Journal}, 9.

\bibitem[{Cozzolino et~al.(2021)Cozzolino, R{\"{o}}ssler, Thies, Nie{\ss}ner,
  and Verdoliva}]{CozzolinoRTNV21}
Davide Cozzolino, Andreas R{\"{o}}ssler, Justus Thies, Matthias Nie{\ss}ner,
  and Luisa Verdoliva. 2021.
\newblock \href {https://doi.org/10.1109/ICCV48922.2021.01483} {Id-reveal:
  Identity-aware deepfake video detection}.
\newblock In \emph{2021 {IEEE/CVF} International Conference on Computer Vision,
  {ICCV} 2021, Montreal, QC, Canada, October 10-17, 2021}, pages 15088--15097.
  {IEEE}.

\bibitem[{Dang et~al.(2020)Dang, Liu, Stehouwer, Liu, and
  Jain}]{DBLP:conf/cvpr/DangLS0020}
Hao Dang, Feng Liu, Joel Stehouwer, Xiaoming Liu, and Anil~K. Jain. 2020.
\newblock \href {https://doi.org/10.1109/CVPR42600.2020.00582} {On the
  detection of digital face manipulation}.
\newblock In \emph{2020 {IEEE/CVF} Conference on Computer Vision and Pattern
  Recognition, {CVPR} 2020, Seattle, WA, USA, June 13-19, 2020}, pages
  5780--5789. Computer Vision Foundation / {IEEE}.

\bibitem[{Dolhansky et~al.(2019)Dolhansky, Howes, Pflaum, Baram, and
  Canton{-}Ferrer}]{Brian2019}
Brian Dolhansky, Russ Howes, Ben Pflaum, Nicole Baram, and Cristian
  Canton{-}Ferrer. 2019.
\newblock \href {http://arxiv.org/abs/1910.08854} {The deepfake detection
  challenge {(DFDC)} preview dataset}.
\newblock \emph{CoRR}, abs/1910.08854.

\bibitem[{Dosovitskiy et~al.(2021)Dosovitskiy, Beyer, Kolesnikov, Weissenborn,
  Zhai, Unterthiner, Dehghani, Minderer, Heigold, Gelly, Uszkoreit, and
  Houlsby}]{DosovitskiyB0WZ21}
Alexey Dosovitskiy, Lucas Beyer, Alexander Kolesnikov, Dirk Weissenborn,
  Xiaohua Zhai, Thomas Unterthiner, Mostafa Dehghani, Matthias Minderer, Georg
  Heigold, Sylvain Gelly, Jakob Uszkoreit, and Neil Houlsby. 2021.
\newblock \href {https://openreview.net/forum?id=YicbFdNTTy} {An image is worth
  16x16 words: Transformers for image recognition at scale}.
\newblock In \emph{9th International Conference on Learning Representations,
  {ICLR} 2021, Virtual Event, Austria, May 3-7, 2021}. OpenReview.net.

\bibitem[{Eyben et~al.(2013)Eyben, Weninger, Gro{\ss}, and
  Schuller}]{DBLP:conf/mm/EybenWGS13}
Florian Eyben, Felix Weninger, Florian Gro{\ss}, and Bj{\"{o}}rn~W. Schuller.
  2013.
\newblock \href {https://doi.org/10.1145/2502081.2502224} {Recent developments
  in opensmile, the munich open-source multimedia feature extractor}.
\newblock In \emph{{ACM} Multimedia Conference, {MM} '13, Barcelona, Spain,
  October 21-25, 2013}, pages 835--838. {ACM}.

\bibitem[{Eyben et~al.(2009)Eyben, W{\"{o}}llmer, and
  Schuller}]{DBLP:conf/acii/EybenWS09}
Florian Eyben, Martin W{\"{o}}llmer, and Bj{\"{o}}rn~W. Schuller. 2009.
\newblock \href {https://doi.org/10.1109/ACII.2009.5349350} {Openear -
  introducing the munich open-source emotion and affect recognition toolkit}.
\newblock In \emph{Affective Computing and Intelligent Interaction, Third
  International Conference and Workshops, {ACII} 2009, Amsterdam, The
  Netherlands, September 10-12, 2009, Proceedings}, pages 1--6. {IEEE} Computer
  Society.

\bibitem[{Fung et~al.(2021)Fung, Thomas, Gangi~Reddy, Polisetty, Ji, Chang,
  McKeown, Bansal, and Sil}]{fung-etal-2021-infosurgeon}
Yi~Fung, Christopher Thomas, Revanth Gangi~Reddy, Sandeep Polisetty, Heng Ji,
  Shih-Fu Chang, Kathleen McKeown, Mohit Bansal, and Avi Sil. 2021.
\newblock \href {https://doi.org/10.18653/v1/2021.acl-long.133}
  {{I}nfo{S}urgeon: Cross-media fine-grained information consistency checking
  for fake news detection}.
\newblock In \emph{Proceedings of the 59th Annual Meeting of the Association
  for Computational Linguistics and the 11th International Joint Conference on
  Natural Language Processing (Volume 1: Long Papers)}, pages 1683--1698,
  Online. Association for Computational Linguistics.

\bibitem[{Garimella and Eckles(2020)}]{Garimella2020}
Kiran Garimella and Dean Eckles. 2020.
\newblock \href {http://arxiv.org/abs/2005.09784} {Images and misinformation in
  political groups: Evidence from whatsapp in india}.
\newblock \emph{CoRR}, abs/2005.09784.

\bibitem[{Goodfellow et~al.(2020)Goodfellow, Pouget{-}Abadie, Mirza, Xu,
  Warde{-}Farley, Ozair, Courville, and Bengio}]{GoodfellowPMXWO20}
Ian~J. Goodfellow, Jean Pouget{-}Abadie, Mehdi Mirza, Bing Xu, David
  Warde{-}Farley, Sherjil Ozair, Aaron~C. Courville, and Yoshua Bengio. 2020.
\newblock \href {https://doi.org/10.1145/3422622} {Generative adversarial
  networks}.
\newblock \emph{Commun. {ACM}}, 63(11):139--144.

\bibitem[{Graves(2018)}]{graves2018understanding}
D~Graves. 2018.
\newblock Understanding the promise and limits of automated fact-checking.

\bibitem[{Guarnera et~al.(2023)Guarnera, Giudice, and
  Battiato}]{GuarneraGB2023}
Luca Guarnera, Oliver Giudice, and Sebastiano Battiato. 2023.
\newblock \href {https://doi.org/10.48550/arXiv.2303.00608} {Level up the
  deepfake detection: a method to effectively discriminate images generated by
  {GAN} architectures and diffusion models}.
\newblock \emph{CoRR}, abs/2303.00608.

\bibitem[{Guera and Delp(2018)}]{GueraD18}
David Guera and Edward~J. Delp. 2018.
\newblock \href {https://doi.org/10.1109/AVSS.2018.8639163} {Deepfake video
  detection using recurrent neural networks}.
\newblock In \emph{15th {IEEE} International Conference on Advanced Video and
  Signal Based Surveillance, {AVSS} 2018, Auckland, New Zealand, November
  27-30, 2018}, pages 1--6. {IEEE}.

\bibitem[{Guo et~al.(2019)Guo, Wang, and Wang}]{DBLP:journals/access/GuoWW19}
Wenzhong Guo, Jianwen Wang, and Shiping Wang. 2019.
\newblock \href {https://doi.org/10.1109/ACCESS.2019.2916887} {Deep multimodal
  representation learning: {A} survey}.
\newblock \emph{{IEEE} Access}, 7:63373--63394.

\bibitem[{Guo et~al.(2022)Guo, Schlichtkrull, and
  Vlachos}]{guo-etal-2022-survey}
Zhijiang Guo, Michael Schlichtkrull, and Andreas Vlachos. 2022.
\newblock \href {https://doi.org/10.1162/tacl_a_00454} {A survey on automated
  fact-checking}.
\newblock \emph{Transactions of the Association for Computational Linguistics},
  10:178--206.

\bibitem[{Gupta et~al.(2013)Gupta, Lamba, Kumaraguru, and Joshi}]{Gupta13}
Aditi Gupta, Hemank Lamba, Ponnurangam Kumaraguru, and Anupam Joshi. 2013.
\newblock \href {https://doi.org/10.1145/2487788.2488033} {Faking sandy:
  characterizing and identifying fake images on twitter during hurricane
  sandy}.
\newblock In \emph{22nd International World Wide Web Conference, {WWW} '13, Rio
  de Janeiro, Brazil, May 13-17, 2013, Companion Volume}, pages 729--736.
  International World Wide Web Conferences Steering Committee / {ACM}.

\bibitem[{Gupta and Srikumar(2021)}]{gupta-srikumar-2021-x}
Ashim Gupta and Vivek Srikumar. 2021.
\newblock \href {https://doi.org/10.18653/v1/2021.acl-short.86} {{X}-fact: A
  new benchmark dataset for multilingual fact checking}.
\newblock In \emph{Proceedings of the 59th Annual Meeting of the Association
  for Computational Linguistics and the 11th International Joint Conference on
  Natural Language Processing (Volume 2: Short Papers)}, pages 675--682,
  Online. Association for Computational Linguistics.

\bibitem[{Gupta et~al.(2022)Gupta, Kumari, Ashok, Ghosal, and
  Ekbal}]{GuptaKAGE22}
Vipin Gupta, Rina Kumari, Nischal Ashok, Tirthankar Ghosal, and Asif Ekbal.
  2022.
\newblock \href {https://aclanthology.org/2022.findings-aacl.43} {{MMM:} an
  emotion and novelty-aware approach for multilingual multimodal misinformation
  detection}.
\newblock In \emph{Findings of the Association for Computational Linguistics:
  {AACL-IJCNLP} 2022, Online only, November 20-23, 2022}, pages 464--477.
  Association for Computational Linguistics.

\bibitem[{Hameleers et~al.(2020)Hameleers, Powell, Van Der~Meer, and
  Bos}]{hameleers2020picture}
Michael Hameleers, Thomas~E Powell, Toni~GLA Van Der~Meer, and Lieke Bos. 2020.
\newblock A picture paints a thousand lies? the effects and mechanisms of
  multimodal disinformation and rebuttals disseminated via social media.
\newblock \emph{Political Communication}, 37(2):281--301.

\bibitem[{Hammouchi and Ghogho(2022)}]{HammouchiG22}
Hicham Hammouchi and Mounir Ghogho. 2022.
\newblock \href {https://doi.org/10.1109/ACCESS.2022.3220690} {Evidence-aware
  multilingual fake news detection}.
\newblock \emph{IEEE Access}, 10:116808--116818.

\bibitem[{Hassan et~al.(2015{\natexlab{a}})Hassan, Adair, Hamilton, Li,
  Tremayne, Yang, and Yu}]{hassan2015quest}
Naeemul Hassan, Bill Adair, James~T Hamilton, Chengkai Li, Mark Tremayne, Jun
  Yang, and Cong Yu. 2015{\natexlab{a}}.
\newblock The quest to automate fact-checking.
\newblock In \emph{Proceedings of the 2015 computation+ journalism symposium}.
  Citeseer.

\bibitem[{Hassan et~al.(2015{\natexlab{b}})Hassan, Li, and
  Tremayne}]{DBLP:conf/cikm/HassanLT15}
Naeemul Hassan, Chengkai Li, and Mark Tremayne. 2015{\natexlab{b}}.
\newblock \href {https://doi.org/10.1145/2806416.2806652} {Detecting
  check-worthy factual claims in presidential debates}.
\newblock In \emph{Proceedings of the 24th {ACM} International Conference on
  Information and Knowledge Management, {CIKM} 2015, Melbourne, VIC, Australia,
  October 19 - 23, 2015}, pages 1835--1838. {ACM}.

\bibitem[{He et~al.(2017)He, Gkioxari, Dollár, and Girshick}]{HeGDG17}
Kaiming He, Georgia Gkioxari, Piotr Dollár, and Ross Girshick. 2017.
\newblock \href {https://doi.org/10.1109/ICCV.2017.322} {Mask r-cnn}.
\newblock In \emph{2017 IEEE International Conference on Computer Vision
  (ICCV)}, pages 2980--2988.

\bibitem[{He et~al.(2016)He, Zhang, Ren, and Sun}]{HeZRS16}
Kaiming He, Xiangyu Zhang, Shaoqing Ren, and Jian Sun. 2016.
\newblock \href {https://doi.org/10.1109/CVPR.2016.90} {Deep residual learning
  for image recognition}.
\newblock In \emph{2016 IEEE Conference on Computer Vision and Pattern
  Recognition (CVPR)}, pages 770--778.

\bibitem[{Heller et~al.(2018)Heller, Rossetto, and Schuldt}]{Heller18}
Silvan Heller, Luca Rossetto, and Heiko Schuldt. 2018.
\newblock \href {http://arxiv.org/abs/1804.04866} {The ps-battles dataset - an
  image collection for image manipulation detection}.
\newblock \emph{CoRR}, abs/1804.04866.

\bibitem[{Hochreiter and Schmidhuber(1997)}]{HochreiterS97}
Sepp Hochreiter and J\"{u}rgen Schmidhuber. 1997.
\newblock \href {https://doi.org/10.1162/neco.1997.9.8.1735} {Long short-term
  memory}.
\newblock \emph{Neural Comput.}, 9(8):1735–1780.

\bibitem[{Hou et~al.(2019)Hou, P{\'{e}}rez{-}Rosas, Loeb, and
  Mihalcea}]{DBLP:conf/icmi/HouPLM19}
Rui Hou, Ver{\'{o}}nica P{\'{e}}rez{-}Rosas, Stacy~L. Loeb, and Rada Mihalcea.
  2019.
\newblock \href {https://doi.org/10.1145/1122445.3353763} {Towards automatic
  detection of misinformation in online medical videos}.
\newblock In \emph{International Conference on Multimodal Interaction, {ICMI}
  2019, Suzhou, China, October 14-18, 2019}, pages 235--243. {ACM}.

\bibitem[{Hu et~al.(2023)Hu, Guo, Chen, Wen, and Yu}]{HuGCWY23}
Xuming Hu, Zhijiang Guo, Junzhe Chen, Lijie Wen, and Philip~S. Yu. 2023.
\newblock \href {https://doi.org/10.1145/3539618.3591896} {{MR2:} {A} benchmark
  for multimodal retrieval-augmented rumor detection in social media}.
\newblock In \emph{Proceedings of the 46th International {ACM} {SIGIR}
  Conference on Research and Development in Information Retrieval, {SIGIR}
  2023, Taipei, Taiwan, July 23-27, 2023}, pages 2901--2912. {ACM}.

\bibitem[{Huh et~al.(2018)Huh, Liu, Owens, and Efros}]{HuhLOE18}
Minyoung Huh, Andrew Liu, Andrew Owens, and Alexei~A. Efros. 2018.
\newblock \href {https://doi.org/10.1007/978-3-030-01252-6\_7} {Fighting fake
  news: Image splice detection via learned self-consistency}.
\newblock In \emph{Computer Vision - {ECCV} 2018 - 15th European Conference,
  Munich, Germany, September 8-14, 2018, Proceedings, Part {XI}}, volume 11215
  of \emph{Lecture Notes in Computer Science}, pages 106--124. Springer.

\bibitem[{Ireton and Posetti(2018)}]{ireton2018journalism}
Cherilyn Ireton and Julie Posetti. 2018.
\newblock \emph{Journalism, fake news \& disinformation: handbook for
  journalism education and training}.
\newblock Unesco Publishing.

\bibitem[{{Ismael Al-Sanjary} et~al.(2016){Ismael Al-Sanjary}, Ahmed, and
  Sulong}]{ISMAELALSANJARY2016565}
Omar {Ismael Al-Sanjary}, Ahmed~Abdullah Ahmed, and Ghazali Sulong. 2016.
\newblock \href
  {https://doi.org/https://doi.org/10.1016/j.forsciint.2016.07.013}
  {Development of a video tampering dataset for forensic investigation}.
\newblock \emph{Forensic Science International}, 266:565--572.

\bibitem[{Jaiswal et~al.(2017)Jaiswal, Sabir, Abd{-}Almageed, and
  Natarajan}]{JaiswalSAN17}
Ayush Jaiswal, Ekraam Sabir, Wael Abd{-}Almageed, and Premkumar Natarajan.
  2017.
\newblock \href {https://doi.org/10.1145/3123266.3123385} {Multimedia semantic
  integrity assessment using joint embedding of images and text}.
\newblock In \emph{Proceedings of the 2017 {ACM} on Multimedia Conference, {MM}
  2017, Mountain View, CA, USA, October 23-27, 2017}, pages 1465--1471. {ACM}.

\bibitem[{Jiang et~al.(2020)Jiang, Li, Wu, Qian, and Loy}]{JiangLW0L20}
Liming Jiang, Ren Li, Wayne Wu, Chen Qian, and Chen~Change Loy. 2020.
\newblock \href {https://doi.org/10.1109/CVPR42600.2020.00296}
  {Deeperforensics-1.0: {A} large-scale dataset for real-world face forgery
  detection}.
\newblock In \emph{2020 {IEEE/CVF} Conference on Computer Vision and Pattern
  Recognition, {CVPR} 2020, Seattle, WA, USA, June 13-19, 2020}, pages
  2886--2895. Computer Vision Foundation / {IEEE}.

\bibitem[{Jin et~al.(2022)Jin, Lalwani, Vaidhya, Shen, Ding, Lyu, Sachan,
  Mihalcea, and Schoelkopf}]{jin-etal-2022-logical}
Zhijing Jin, Abhinav Lalwani, Tejas Vaidhya, Xiaoyu Shen, Yiwen Ding, Zhiheng
  Lyu, Mrinmaya Sachan, Rada Mihalcea, and Bernhard Schoelkopf. 2022.
\newblock \href {https://aclanthology.org/2022.findings-emnlp.532} {Logical
  fallacy detection}.
\newblock In \emph{Findings of the Association for Computational Linguistics:
  EMNLP 2022}, pages 7180--7198, Abu Dhabi, United Arab Emirates. Association
  for Computational Linguistics.

\bibitem[{Jin et~al.(2017)Jin, Cao, Guo, Zhang, and Luo}]{JinCGZL17}
Zhiwei Jin, Juan Cao, Han Guo, Yongdong Zhang, and Jiebo Luo. 2017.
\newblock \href {https://doi.org/10.1145/3123266.3123454} {Multimodal fusion
  with recurrent neural networks for rumor detection on microblogs}.
\newblock In \emph{Proceedings of the 2017 {ACM} on Multimedia Conference, {MM}
  2017, Mountain View, CA, USA, October 23-27, 2017}, pages 795--816. {ACM}.

\bibitem[{Kalb and Saivetz(2007)}]{KalbC07}
Marvin Kalb and Carol Saivetz. 2007.
\newblock \href {https://doi.org/10.1177/1081180X07303934} {The
  israeli—hezbollah war of 2006: The media as a weapon in asymmetrical
  conflict}.
\newblock \emph{Harvard International Journal of Press/Politics}, 12(3):43--66.

\bibitem[{Kamboj et~al.(2021)Kamboj, Hessler, Asnani, Riani, and
  Abouelenien}]{DBLP:journals/ieeemm/KambojHARA21}
Manvi Kamboj, Christian Hessler, Priyanka Asnani, Kais Riani, and Mohamed
  Abouelenien. 2021.
\newblock \href {https://doi.org/10.1109/MMUL.2020.3048044} {Multimodal
  political deception detection}.
\newblock \emph{{IEEE} Multim.}, 28(1):94--102.

\bibitem[{Karras et~al.(2019)Karras, Laine, and Aila}]{KarrasLA19}
Tero Karras, Samuli Laine, and Timo Aila. 2019.
\newblock \href {https://doi.org/10.1109/CVPR.2019.00453} {A style-based
  generator architecture for generative adversarial networks}.
\newblock In \emph{{IEEE} Conference on Computer Vision and Pattern
  Recognition, {CVPR} 2019, Long Beach, CA, USA, June 16-20, 2019}, pages
  4401--4410. Computer Vision Foundation / {IEEE}.

\bibitem[{Khalid et~al.(2021)Khalid, Tariq, Kim, and
  Woo}]{DBLP:conf/nips/KhalidTKW21}
Hasam Khalid, Shahroz Tariq, Minha Kim, and Simon~S. Woo. 2021.
\newblock \href
  {https://datasets-benchmarks-proceedings.neurips.cc/paper/2021/hash/d9d4f495e875a2e075a1a4a6e1b9770f-Abstract-round2.html}
  {Fakeavceleb: {A} novel audio-video multimodal deepfake dataset}.
\newblock In \emph{Proceedings of the Neural Information Processing Systems
  Track on Datasets and Benchmarks 1, NeurIPS Datasets and Benchmarks 2021,
  December 2021, virtual}.

\bibitem[{Khattar et~al.(2019)Khattar, Goud, Gupta, and Varma}]{KhattarG0V19}
Dhruv Khattar, Jaipal~Singh Goud, Manish Gupta, and Vasudeva Varma. 2019.
\newblock \href {https://doi.org/10.1145/3308558.3313552} {{MVAE:} multimodal
  variational autoencoder for fake news detection}.
\newblock In \emph{The World Wide Web Conference, {WWW} 2019, San Francisco,
  CA, USA, May 13-17, 2019}, pages 2915--2921. {ACM}.

\bibitem[{Kingma and Welling(2014)}]{KingmaW13}
Diederik~P. Kingma and Max Welling. 2014.
\newblock \href {http://arxiv.org/abs/1312.6114} {Auto-encoding variational
  bayes}.
\newblock In \emph{2nd International Conference on Learning Representations,
  {ICLR} 2014, Banff, AB, Canada, April 14-16, 2014, Conference Track
  Proceedings}.

\bibitem[{Kinnunen et~al.(2017)Kinnunen, Sahidullah, Delgado, Todisco, Evans,
  Yamagishi, and Lee}]{KinnunenSDTEYL17}
Tomi Kinnunen, Md. Sahidullah, H{\'{e}}ctor Delgado, Massimiliano Todisco,
  Nicholas W.~D. Evans, Junichi Yamagishi, and Kong{-}Aik Lee. 2017.
\newblock \href
  {http://www.isca-speech.org/archive/Interspeech\_2017/abstracts/1111.html}
  {The asvspoof 2017 challenge: Assessing the limits of replay spoofing attack
  detection}.
\newblock In \emph{Interspeech 2017, 18th Annual Conference of the
  International Speech Communication Association, Stockholm, Sweden, August
  20-24, 2017}, pages 2--6. {ISCA}.

\bibitem[{Kipf and Welling(2017)}]{KipfW17}
Thomas~N. Kipf and Max Welling. 2017.
\newblock \href {https://openreview.net/forum?id=SJU4ayYgl} {Semi-supervised
  classification with graph convolutional networks}.
\newblock In \emph{5th International Conference on Learning Representations,
  {ICLR} 2017, Toulon, France, April 24-26, 2017, Conference Track
  Proceedings}. OpenReview.net.

\bibitem[{Kopev et~al.(2019)Kopev, Ali, Koychev, and
  Nakov}]{DBLP:conf/asru/KopevAKN19}
Daniel Kopev, Ahmed Ali, Ivan Koychev, and Preslav Nakov. 2019.
\newblock \href {https://doi.org/10.1109/ASRU46091.2019.9003892} {Detecting
  deception in political debates using acoustic and textual features}.
\newblock In \emph{{IEEE} Automatic Speech Recognition and Understanding
  Workshop, {ASRU} 2019, Singapore, December 14-18, 2019}, pages 652--659.
  {IEEE}.

\bibitem[{Kotonya and Toni(2020{\natexlab{a}})}]{kotonya-toni-2020-explainable}
Neema Kotonya and Francesca Toni. 2020{\natexlab{a}}.
\newblock \href {https://doi.org/10.18653/v1/2020.coling-main.474} {Explainable
  automated fact-checking: A survey}.
\newblock In \emph{Proceedings of the 28th International Conference on
  Computational Linguistics}, pages 5430--5443, Barcelona, Spain (Online).
  International Committee on Computational Linguistics.

\bibitem[{Kotonya and
  Toni(2020{\natexlab{b}})}]{kotonya-toni-2020-explainable-automated}
Neema Kotonya and Francesca Toni. 2020{\natexlab{b}}.
\newblock \href {https://doi.org/10.18653/v1/2020.emnlp-main.623} {Explainable
  automated fact-checking for public health claims}.
\newblock In \emph{Proceedings of the 2020 Conference on Empirical Methods in
  Natural Language Processing (EMNLP)}, pages 7740--7754, Online. Association
  for Computational Linguistics.

\bibitem[{Kou et~al.(2020)Kou, Zhang, Shang, and Wang}]{KouZSW20}
Ziyi Kou, Daniel~Yue Zhang, Lanyu Shang, and Dong Wang. 2020.
\newblock \href {https://doi.org/10.1109/BigData50022.2020.9378019} {Exfaux:
  {A} weakly supervised approach to explainable fauxtography detection}.
\newblock In \emph{2020 {IEEE} International Conference on Big Data {(IEEE}
  BigData 2020), Atlanta, GA, USA, December 10-13, 2020}, pages 631--636.
  {IEEE}.

\bibitem[{Kwon et~al.(2021)Kwon, You, Nam, Park, and
  Chae}]{DBLP:conf/iccv/KwonYNPC21}
Patrick Kwon, Jaeseong You, Gyuhyeon Nam, Sungwoo Park, and Gyeongsu Chae.
  2021.
\newblock \href {https://doi.org/10.1109/ICCV48922.2021.01057} {Kodf: {A}
  large-scale korean deepfake detection dataset}.
\newblock In \emph{2021 {IEEE/CVF} International Conference on Computer Vision,
  {ICCV} 2021, Montreal, QC, Canada, October 10-17, 2021}, pages 10724--10733.
  {IEEE}.

\bibitem[{La et~al.(2022)La, Tran, Tran, Tran, Dang{-}Nguyen, and
  Dao}]{DBLP:conf/mir/LaTTTDD22}
Tuan{-}Vinh La, Quang{-}Tien Tran, Thanh{-}Phuc Tran, Anh{-}Duy Tran,
  Duc{-}Tien Dang{-}Nguyen, and Minh{-}Son Dao. 2022.
\newblock \href {https://doi.org/10.1145/3512731.3534210} {Multimodal
  cheapfakes detection by utilizing image captioning for global context}.
\newblock In \emph{ICDAR@ICMR 2022: Proceedings of the 3rd {ACM} Workshop on
  Intelligent Cross-Data Analysis and Retrieval, Newark, NJ, USA, June 27 - 30,
  2022}, pages 9--16. {ACM}.

\bibitem[{Lauer(2009)}]{LAUER2009225}
Claire Lauer. 2009.
\newblock \href {https://doi.org/https://doi.org/10.1016/j.compcom.2009.09.001}
  {Contending with terms: “multimodal” and “multimedia” in the academic
  and public spheres}.
\newblock \emph{Computers and Composition}, 26(4):225--239.

\bibitem[{Lavrentyeva et~al.(2019)Lavrentyeva, Novoselov, Volkova, Matveev, and
  Marsico}]{LavrentyevaNVMM19}
Galina Lavrentyeva, Sergey Novoselov, Marina Volkova, Yuri Matveev, and
  Maria~De Marsico. 2019.
\newblock \href {https://doi.org/10.1109/ICASSP.2019.8682942} {Phonespoof: {A}
  new dataset for spoofing attack detection in telephone channel}.
\newblock In \emph{{IEEE} International Conference on Acoustics, Speech and
  Signal Processing, {ICASSP} 2019, Brighton, United Kingdom, May 12-17, 2019},
  pages 2572--2576. {IEEE}.

\bibitem[{Lee et~al.(2022)Lee, Joshi, Turc, Hu, Liu, Eisenschlos, Khandelwal,
  Shaw, Chang, and Toutanova}]{LeeJTHLEKSCT2022}
Kenton Lee, Mandar Joshi, Iulia Turc, Hexiang Hu, Fangyu Liu, Julian
  Eisenschlos, Urvashi Khandelwal, Peter Shaw, Ming{-}Wei Chang, and Kristina
  Toutanova. 2022.
\newblock \href {https://doi.org/10.48550/arXiv.2210.03347} {Pix2struct:
  Screenshot parsing as pretraining for visual language understanding}.
\newblock \emph{CoRR}, abs/2210.03347.

\bibitem[{Lewandowsky et~al.(2020)Lewandowsky, Cook, and
  Lombardi}]{lewandowsky_debunking_2020}
Stephan Lewandowsky, John Cook, and Doug Lombardi. 2020.
\newblock \href {https://doi.org/10.17910/B7.1182} {Debunking {Handbook} 2020}.

\bibitem[{Li et~al.(2020{\natexlab{a}})Li, Zareian, Zeng, Whitehead, Lu, Ji,
  and Chang}]{li-etal-2020-cross}
Manling Li, Alireza Zareian, Qi~Zeng, Spencer Whitehead, Di~Lu, Heng Ji, and
  Shih-Fu Chang. 2020{\natexlab{a}}.
\newblock \href {https://doi.org/10.18653/v1/2020.acl-main.230} {Cross-media
  structured common space for multimedia event extraction}.
\newblock In \emph{Proceedings of the 58th Annual Meeting of the Association
  for Computational Linguistics}, pages 2557--2568, Online. Association for
  Computational Linguistics.

\bibitem[{Li and Xie(2020)}]{li2020picture}
Yiyi Li and Ying Xie. 2020.
\newblock Is a picture worth a thousand words? an empirical study of image
  content and social media engagement.
\newblock \emph{Journal of Marketing Research}, 57(1):1--19.

\bibitem[{Li et~al.(2020{\natexlab{b}})Li, Yang, Sun, Qi, and Lyu}]{LiYSQL20}
Yuezun Li, Xin Yang, Pu~Sun, Honggang Qi, and Siwei Lyu. 2020{\natexlab{b}}.
\newblock \href {https://doi.org/10.1109/CVPR42600.2020.00327} {Celeb-df: {A}
  large-scale challenging dataset for deepfake forensics}.
\newblock In \emph{2020 {IEEE/CVF} Conference on Computer Vision and Pattern
  Recognition, {CVPR} 2020, Seattle, WA, USA, June 13-19, 2020}, pages
  3204--3213. Computer Vision Foundation / {IEEE}.

\bibitem[{Liang et~al.(2022)Liang, Zadeh, and
  Morency}]{DBLP:journals/corr/LiangZM22}
Paul~Pu Liang, Amir Zadeh, and Louis{-}Philippe Morency. 2022.
\newblock \href {https://doi.org/10.48550/arXiv.2209.03430} {Foundations and
  recent trends in multimodal machine learning: Principles, challenges, and
  open questions}.
\newblock \emph{CoRR}, abs/2209.03430.

\bibitem[{Lin et~al.(2020)Lin, Ji, Huang, and Wu}]{lin-etal-2020-joint}
Ying Lin, Heng Ji, Fei Huang, and Lingfei Wu. 2020.
\newblock \href {https://doi.org/10.18653/v1/2020.acl-main.713} {A joint neural
  model for information extraction with global features}.
\newblock In \emph{Proceedings of the 58th Annual Meeting of the Association
  for Computational Linguistics}, pages 7999--8009, Online. Association for
  Computational Linguistics.

\bibitem[{Liu et~al.(2021)Liu, Bugliarello, Ponti, Reddy, Collier, and
  Elliott}]{liu-etal-2021-visually}
Fangyu Liu, Emanuele Bugliarello, Edoardo~Maria Ponti, Siva Reddy, Nigel
  Collier, and Desmond Elliott. 2021.
\newblock \href {https://doi.org/10.18653/v1/2021.emnlp-main.818} {Visually
  grounded reasoning across languages and cultures}.
\newblock In \emph{Proceedings of the 2021 Conference on Empirical Methods in
  Natural Language Processing}, pages 10467--10485, Online and Punta Cana,
  Dominican Republic. Association for Computational Linguistics.

\bibitem[{Liu et~al.(2023{\natexlab{a}})Liu, Yacoob, and Shrivastava}]{LiuYS23}
Fuxiao Liu, Yaser Yacoob, and Abhinav Shrivastava. 2023{\natexlab{a}}.
\newblock \href {https://aclanthology.org/2023.eacl-main.14} {{COVID-VTS:} fact
  extraction and verification on short video platforms}.
\newblock In \emph{Proceedings of the 17th Conference of the European Chapter
  of the Association for Computational Linguistics, {EACL} 2023, Dubrovnik,
  Croatia, May 2-6, 2023}, pages 178--188. Association for Computational
  Linguistics.

\bibitem[{Liu et~al.(2023{\natexlab{b}})Liu, Wang, and Li}]{LiuWL23}
Hui Liu, Wenya Wang, and Haoliang Li. 2023{\natexlab{b}}.
\newblock \href {https://doi.org/10.18653/v1/2023.findings-acl.620}
  {Interpretable multimodal misinformation detection with logic reasoning}.
\newblock In \emph{Findings of the Association for Computational Linguistics:
  {ACL} 2023, Toronto, Canada, July 9-14, 2023}, pages 9781--9796. Association
  for Computational Linguistics.

\bibitem[{Louren{\c{c}}o and Paes(2022)}]{LourencoP22}
V{\'{\i}}tor Louren{\c{c}}o and Aline Paes. 2022.
\newblock \href {https://doi.org/10.48550/arXiv.2212.04272} {A modality-level
  explainable framework for misinformation checking in social networks}.
\newblock \emph{CoRR}, abs/2212.04272.

\bibitem[{Luo et~al.(2021)Luo, Darrell, and
  Rohrbach}]{luo-etal-2021-newsclippings}
Grace Luo, Trevor Darrell, and Anna Rohrbach. 2021.
\newblock \href {https://doi.org/10.18653/v1/2021.emnlp-main.545}
  {{N}ews{CLIP}pings: {A}utomatic {G}eneration of {O}ut-of-{C}ontext
  {M}ultimodal {M}edia}.
\newblock In \emph{Proceedings of the 2021 Conference on Empirical Methods in
  Natural Language Processing}, pages 6801--6817, Online and Punta Cana,
  Dominican Republic. Association for Computational Linguistics.

\bibitem[{Maras and Alexandrou(2019)}]{MarasA19}
Marie-Helen Maras and Alex Alexandrou. 2019.
\newblock \href {https://doi.org/10.1177/1365712718807226} {Determining
  authenticity of video evidence in the age of artificial intelligence and in
  the wake of deepfake videos}.
\newblock \emph{The International Journal of Evidence \& Proof},
  23(3):255--262.

\bibitem[{Maros et~al.(2021)Maros, Almeida, and
  Vasconcelos}]{DBLP:conf/misdoom/MarosAV21}
Alexandre Maros, Jussara~M. Almeida, and Marisa Vasconcelos. 2021.
\newblock \href {https://doi.org/10.1007/978-3-030-87031-7\_6} {A study of
  misinformation in audio messages shared in whatsapp groups}.
\newblock In \emph{Disinformation in Open Online Media - Third
  Multidisciplinary International Symposium, {MISDOOM} 2021, Virtual Event,
  September 21-22, 2021, Proceedings}, volume 12887 of \emph{Lecture Notes in
  Computer Science}, pages 85--100. Springer.

\bibitem[{Meel and Vishwakarma(2021)}]{MEEL202123}
Priyanka Meel and Dinesh~Kumar Vishwakarma. 2021.
\newblock \href {https://doi.org/https://doi.org/10.1016/j.ins.2021.03.037}
  {Han, image captioning, and forensics ensemble multimodal fake news
  detection}.
\newblock \emph{Information Sciences}, 567:23--41.

\bibitem[{Mikolov et~al.(2013)Mikolov, Chen, Corrado, and Dean}]{MikolovCCD13}
Tom{\'{a}}s Mikolov, Kai Chen, Greg Corrado, and Jeffrey Dean. 2013.
\newblock \href {http://arxiv.org/abs/1301.3781} {Efficient estimation of word
  representations in vector space}.
\newblock In \emph{1st International Conference on Learning Representations,
  {ICLR} 2013, Scottsdale, Arizona, USA, May 2-4, 2013, Workshop Track
  Proceedings}.

\bibitem[{Mishra et~al.(2022)Mishra, S, Bhaskar, Chopra, Reganti, Patwa, Das,
  Chakraborty, Sheth, and Ekbal}]{mishra2022factify}
Shreyash Mishra, Suryavardan S, Amrit Bhaskar, Parul Chopra, Aishwarya~N.
  Reganti, Parth Patwa, Amitava Das, Tanmoy Chakraborty, Amit~P. Sheth, and
  Asif Ekbal. 2022.
\newblock \href {http://ceur-ws.org/Vol-3199/paper18.pdf} {{FACTIFY:} {A}
  multi-modal fact verification dataset}.
\newblock In \emph{Proceedings of the Workshop on Multi-Modal Fake News and
  Hate-Speech Detection {(DE-FACTIFY} 2022) co-located with the Thirty-Sixth
  {AAAI} Conference on Artificial Intelligence {(} {AAAI} 2022), Virtual Event,
  Vancouver, Canada, February 27, 2022}, volume 3199 of \emph{{CEUR} Workshop
  Proceedings}. CEUR-WS.org.

\bibitem[{M{\"{u}}ller{-}Budack et~al.(2020)M{\"{u}}ller{-}Budack, Theiner,
  Diering, Idahl, and Ewerth}]{Muller-BudackTD20}
Eric M{\"{u}}ller{-}Budack, Jonas Theiner, Sebastian Diering, Maximilian Idahl,
  and Ralph Ewerth. 2020.
\newblock \href {https://doi.org/10.1145/3372278.3390670} {Multimodal analytics
  for real-world news using measures of cross-modal entity consistency}.
\newblock In \emph{Proceedings of the 2020 on International Conference on
  Multimedia Retrieval, {ICMR} 2020, Dublin, Ireland, June 8-11, 2020}, pages
  16--25. {ACM}.

\bibitem[{Nakamura et~al.(2020)Nakamura, Levy, and Wang}]{NakamuraLW20}
Kai Nakamura, Sharon Levy, and William~Yang Wang. 2020.
\newblock \href {https://aclanthology.org/2020.lrec-1.755/} {Fakeddit: {A} new
  multimodal benchmark dataset for fine-grained fake news detection}.
\newblock In \emph{Proceedings of The 12th Language Resources and Evaluation
  Conference, {LREC} 2020, Marseille, France, May 11-16, 2020}, pages
  6149--6157. European Language Resources Association.

\bibitem[{Nakov et~al.(2018)Nakov, Barr{\'{o}}n{-}Cede{\~{n}}o, Elsayed,
  Suwaileh, M{\`{a}}rquez, Zaghouani, Atanasova, Kyuchukov, and
  Martino}]{NakovBESMZAKM18}
Preslav Nakov, Alberto Barr{\'{o}}n{-}Cede{\~{n}}o, Tamer Elsayed, Reem
  Suwaileh, Llu{\'{\i}}s M{\`{a}}rquez, Wajdi Zaghouani, Pepa Atanasova, Spas
  Kyuchukov, and Giovanni Da~San Martino. 2018.
\newblock \href {https://doi.org/10.1007/978-3-319-98932-7\_32} {Overview of
  the {CLEF-2018} checkthat! lab on automatic identification and verification
  of political claims}.
\newblock In \emph{Experimental {IR} Meets Multilinguality, Multimodality, and
  Interaction - 9th International Conference of the {CLEF} Association, {CLEF}
  2018, Avignon, France, September 10-14, 2018, Proceedings}, volume 11018 of
  \emph{Lecture Notes in Computer Science}, pages 372--387. Springer.

\bibitem[{Nakov et~al.(2021)Nakov, Corney, Hasanain, Alam, Elsayed,
  Barr{\'{o}}n{-}Cede{\~{n}}o, Papotti, Shaar, and Martino}]{NakovCHAEBPSM21}
Preslav Nakov, David P.~A. Corney, Maram Hasanain, Firoj Alam, Tamer Elsayed,
  Alberto Barr{\'{o}}n{-}Cede{\~{n}}o, Paolo Papotti, Shaden Shaar, and
  Giovanni Da~San Martino. 2021.
\newblock \href {https://doi.org/10.24963/ijcai.2021/619} {Automated
  fact-checking for assisting human fact-checkers}.
\newblock In \emph{Proceedings of the Thirtieth International Joint Conference
  on Artificial Intelligence, {IJCAI} 2021, Virtual Event / Montreal, Canada,
  19-27 August 2021}, pages 4551--4558. ijcai.org.

\bibitem[{Narayan et~al.(2023)Narayan, Agarwal, Thakral, Mittal, Vatsa, and
  Singh}]{Narayan_2023_CVPR}
Kartik Narayan, Harsh Agarwal, Kartik Thakral, Surbhi Mittal, Mayank Vatsa, and
  Richa Singh. 2023.
\newblock Df-platter: Multi-face heterogeneous deepfake dataset.
\newblock In \emph{Proceedings of the IEEE/CVF Conference on Computer Vision
  and Pattern Recognition (CVPR)}, pages 9739--9748.

\bibitem[{Newman and Zhang(2020)}]{inbook}
Eryn Newman and Lynn Zhang. 2020.
\newblock \href {https://doi.org/10.4324/9780429295379-8} {\emph{Truthiness:
  How Non-Probative Photos Shape Belief}}.

\bibitem[{Newman et~al.(2012)Newman, Garry, Bernstein, Kantner, and
  Lindsay}]{newman2012nonprobative}
Eryn~J Newman, Maryanne Garry, Daniel~M Bernstein, Justin Kantner, and
  D~Stephen Lindsay. 2012.
\newblock Nonprobative photographs (or words) inflate truthiness.
\newblock \emph{Psychonomic Bulletin \& Review}, 19(5):969--974.

\bibitem[{Newman et~al.(2003)Newman, Pennebaker, Berry, and
  Richards}]{newman2003lying}
Matthew~L Newman, James~W Pennebaker, Diane~S Berry, and Jane~M Richards. 2003.
\newblock Lying words: Predicting deception from linguistic styles.
\newblock \emph{Personality and social psychology bulletin}, 29(5):665--675.

\bibitem[{Nielsen and McConville(2022)}]{NielsenM22}
Dan~Saattrup Nielsen and Ryan McConville. 2022.
\newblock \href {https://doi.org/10.1145/3477495.3531744} {Mumin: {A}
  large-scale multilingual multimodal fact-checked misinformation social
  network dataset}.
\newblock In \emph{{SIGIR} '22: The 45th International {ACM} {SIGIR} Conference
  on Research and Development in Information Retrieval, Madrid, Spain, July 11
  - 15, 2022}, pages 3141--3153. {ACM}.

\bibitem[{Nirkin et~al.(2019)Nirkin, Keller, and Hassner}]{NirkinKH19}
Yuval Nirkin, Yosi Keller, and Tal Hassner. 2019.
\newblock \href {https://doi.org/10.1109/ICCV.2019.00728} {{FSGAN:} subject
  agnostic face swapping and reenactment}.
\newblock In \emph{2019 {IEEE/CVF} International Conference on Computer Vision,
  {ICCV} 2019, Seoul, Korea (South), October 27 - November 2, 2019}, pages
  7183--7192. {IEEE}.

\bibitem[{Paris and Donovan(2019)}]{paris2019deepfakes}
Britt Paris and Joan Donovan. 2019.
\newblock Deepfakes and cheap fakes.
\newblock \emph{United States of America: Data \& Society}, 1.

\bibitem[{Patwa et~al.(2022)Patwa, Mishra, S, Bhaskar, Chopra, Reganti, Das,
  Chakraborty, Sheth, Ekbal, and Ahuja}]{PatwaMSBCRDCSEA22}
Parth Patwa, Shreyash Mishra, Suryavardan S, Amrit Bhaskar, Parul Chopra,
  Aishwarya Reganti, Amitava Das, Tanmoy Chakraborty, Amit Sheth, Asif Ekbal,
  and Chaitanya Ahuja. 2022.
\newblock Benchmarking multi-modal entailment for fact verification.

\bibitem[{P{\'{e}}rez{-}Rosas et~al.(2015)P{\'{e}}rez{-}Rosas, Abouelenien,
  Mihalcea, and Burzo}]{Perez-RosasAMB15}
Ver{\'{o}}nica P{\'{e}}rez{-}Rosas, Mohamed Abouelenien, Rada Mihalcea, and
  Mihai Burzo. 2015.
\newblock \href {https://doi.org/10.1145/2818346.2820758} {Deception detection
  using real-life trial data}.
\newblock In \emph{Proceedings of the 2015 {ACM} on International Conference on
  Multimodal Interaction, Seattle, WA, USA, November 09 - 13, 2015}, pages
  59--66. {ACM}.

\bibitem[{Prabhakar et~al.(2021)Prabhakar, Gupta, Nadig, and
  George}]{PrabhakarGNG21}
Tarunima Prabhakar, Anushree Gupta, Kruttika Nadig, and Denny George. 2021.
\newblock \href {https://ojs.aaai.org/index.php/ICWSM/article/view/18126}
  {Check mate: Prioritizing user generated multi-media content for
  fact-checking}.
\newblock In \emph{Proceedings of the Fifteenth International {AAAI} Conference
  on Web and Social Media, {ICWSM} 2021, held virtually, June 7-10, 2021},
  pages 1025--1033. {AAAI} Press.

\bibitem[{Purwanto et~al.(2021)Purwanto, Santoso, Lei, Yang, and
  Peng}]{DBLP:conf/taai/PurwantoSLYP21}
Christian~Nathaniel Purwanto, Joan Santoso, Po{-}Ruey Lei, Hui{-}Kuo Yang, and
  Wen{-}Chih Peng. 2021.
\newblock \href {https://doi.org/10.1109/TAAI54685.2021.00010} {Fakeclip:
  Multimodal fake caption detection with mixed languages for explainable
  visualization}.
\newblock In \emph{2021 International Conference on Technologies and
  Applications of Artificial Intelligence, {TAAI} 2021, Taichung, Taiwan,
  November 18-20, 2021}, pages 1--6. {IEEE}.

\bibitem[{Qi et~al.(2022)Qi, Bu, Cao, Ji, Shui, Xiao, Wang, and Chua}]{qi2023}
Peng Qi, Yuyan Bu, Juan Cao, Wei Ji, Ruihao Shui, Junbin Xiao, Danding Wang,
  and Tat{-}Seng Chua. 2022.
\newblock \href {https://doi.org/10.48550/arXiv.2211.10973} {Fakesv: {A}
  multimodal benchmark with rich social context for fake news detection on
  short video platforms}.
\newblock \emph{CoRR}, abs/2211.10973.

\bibitem[{Qi et~al.(2019)Qi, Cao, Yang, Guo, and Li}]{DBLP:conf/icdm/QiCYGL19}
Peng Qi, Juan Cao, Tianyun Yang, Junbo Guo, and Jintao Li. 2019.
\newblock \href {https://doi.org/10.1109/ICDM.2019.00062} {Exploiting
  multi-domain visual information for fake news detection}.
\newblock In \emph{2019 {IEEE} International Conference on Data Mining, {ICDM}
  2019, Beijing, China, November 8-11, 2019}, pages 518--527. {IEEE}.

\bibitem[{Qi et~al.(2023)Qi, Zhao, Shen, Ji, Cao, and Chua}]{Qi23}
Peng Qi, Yuyang Zhao, Yufeng Shen, Wei Ji, Juan Cao, and Tat{-}Seng Chua. 2023.
\newblock \href {https://doi.org/10.18653/v1/2023.findings-acl.756} {Two heads
  are better than one: Improving fake news video detection by correlating with
  neighbors}.
\newblock In \emph{Findings of the Association for Computational Linguistics:
  {ACL} 2023, Toronto, Canada, July 9-14, 2023}, pages 11947--11959.
  Association for Computational Linguistics.

\bibitem[{Qian et~al.(2021)Qian, Wang, Hu, Fang, and Xu}]{qian2021hierarchical}
Shengsheng Qian, Jinguang Wang, Jun Hu, Quan Fang, and Changsheng Xu. 2021.
\newblock Hierarchical multi-modal contextual attention network for fake news
  detection.
\newblock In \emph{Proceedings of the 44th International ACM SIGIR Conference
  on Research and Development in Information Retrieval}, pages 153--162.

\bibitem[{Qu et~al.(2022{\natexlab{a}})Qu, Li, Zhao, Dev, and Chang}]{qu2022}
Jingnong Qu, Liunian~Harold Li, Jieyu Zhao, Sunipa Dev, and Kai{-}Wei Chang.
  2022{\natexlab{a}}.
\newblock \href {https://doi.org/10.48550/arXiv.2205.12617} {Disinfomeme: {A}
  multimodal dataset for detecting meme intentionally spreading out
  disinformation}.
\newblock \emph{CoRR}, abs/2205.12617.

\bibitem[{Qu et~al.(2022{\natexlab{b}})Qu, Li, Zhao, Dev, and
  Chang}]{DBLP:journals/corr/abs-2205-12617}
Jingnong Qu, Liunian~Harold Li, Jieyu Zhao, Sunipa Dev, and Kai{-}Wei Chang.
  2022{\natexlab{b}}.
\newblock \href {https://doi.org/10.48550/arXiv.2205.12617} {Disinfomeme: {A}
  multimodal dataset for detecting meme intentionally spreading out
  disinformation}.
\newblock \emph{CoRR}, abs/2205.12617.

\bibitem[{Radford et~al.(2021)Radford, Kim, Hallacy, Ramesh, Goh, Agarwal,
  Sastry, Askell, Mishkin, Clark, Krueger, and
  Sutskever}]{DBLP:conf/icml/RadfordKHRGASAM21}
Alec Radford, Jong~Wook Kim, Chris Hallacy, Aditya Ramesh, Gabriel Goh,
  Sandhini Agarwal, Girish Sastry, Amanda Askell, Pamela Mishkin, Jack Clark,
  Gretchen Krueger, and Ilya Sutskever. 2021.
\newblock \href {http://proceedings.mlr.press/v139/radford21a.html} {Learning
  transferable visual models from natural language supervision}.
\newblock In \emph{Proceedings of the 38th International Conference on Machine
  Learning, {ICML} 2021, 18-24 July 2021, Virtual Event}, volume 139 of
  \emph{Proceedings of Machine Learning Research}, pages 8748--8763. {PMLR}.

\bibitem[{Reimao and Tzerpos(2019)}]{ReimaoT19}
Ricardo Reimao and Vassilios Tzerpos. 2019.
\newblock \href {https://doi.org/10.1109/SPED.2019.8906599} {For: {A} dataset
  for synthetic speech detection}.
\newblock In \emph{2019 International Conference on Speech Technology and
  Human-Computer Dialogue, SpeD 2019, Timisoara, Romania, October 10-12, 2019},
  pages 1--10. {IEEE}.

\bibitem[{Resende et~al.(2019)Resende, Melo, Sousa, Messias, Vasconcelos,
  Almeida, and Benevenuto}]{ResendeMSMVAB19}
Gustavo Resende, Philipe~F. Melo, Hugo Sousa, Johnnatan Messias, Marisa
  Vasconcelos, Jussara~M. Almeida, and Fabr{\'{\i}}cio Benevenuto. 2019.
\newblock \href {https://doi.org/10.1145/3308558.3313688} {(mis)information
  dissemination in whatsapp: Gathering, analyzing and countermeasures}.
\newblock In \emph{The World Wide Web Conference, {WWW} 2019, San Francisco,
  CA, USA, May 13-17, 2019}, pages 818--828. {ACM}.

\bibitem[{Ricker et~al.(2022)Ricker, Damm, Holz, and
  Fischer}]{ricker-etal-2022-towards}
Jonas Ricker, Simon Damm, Thorsten Holz, and Asja Fischer. 2022.
\newblock \href {https://doi.org/10.48550/arXiv.2210.14571} {Towards the
  detection of diffusion model deepfakes}.
\newblock \emph{CoRR}, abs/2210.14571.

\bibitem[{Rombach et~al.(2022)Rombach, Blattmann, Lorenz, Esser, and
  Ommer}]{Rombach_2022_CVPR}
Robin Rombach, Andreas Blattmann, Dominik Lorenz, Patrick Esser, and Bj\"orn
  Ommer. 2022.
\newblock High-resolution image synthesis with latent diffusion models.
\newblock In \emph{Proceedings of the IEEE/CVF Conference on Computer Vision
  and Pattern Recognition (CVPR)}, pages 10684--10695.

\bibitem[{R{\"{o}}ssler et~al.(2018)R{\"{o}}ssler, Cozzolino, Verdoliva, Riess,
  Thies, and Nie{\ss}ner}]{Andreas18}
Andreas R{\"{o}}ssler, Davide Cozzolino, Luisa Verdoliva, Christian Riess,
  Justus Thies, and Matthias Nie{\ss}ner. 2018.
\newblock \href {http://arxiv.org/abs/1803.09179} {Faceforensics: {A}
  large-scale video dataset for forgery detection in human faces}.
\newblock \emph{CoRR}, abs/1803.09179.

\bibitem[{Roy and Ekbal(2021)}]{DBLP:conf/ijcnn/RoyE21}
Arjun Roy and Asif Ekbal. 2021.
\newblock \href {https://doi.org/10.1109/IJCNN52387.2021.9533916}
  {Mulcob-mulfav: Multimodal content based multilingual fact verification}.
\newblock In \emph{International Joint Conference on Neural Networks, {IJCNN}
  2021, Shenzhen, China, July 18-22, 2021}, pages 1--8. {IEEE}.

\bibitem[{Ruder et~al.(2022)Ruder, Vuli{\'c}, and
  S{\o}gaard}]{ruder-etal-2022-square}
Sebastian Ruder, Ivan Vuli{\'c}, and Anders S{\o}gaard. 2022.
\newblock \href {https://doi.org/10.18653/v1/2022.findings-acl.184} {Square one
  bias in {NLP}: Towards a multi-dimensional exploration of the research
  manifold}.
\newblock In \emph{Findings of the Association for Computational Linguistics:
  ACL 2022}, pages 2340--2354, Dublin, Ireland. Association for Computational
  Linguistics.

\bibitem[{Sabir et~al.(2018)Sabir, AbdAlmageed, Wu, and Natarajan}]{SabirA0N18}
Ekraam Sabir, Wael AbdAlmageed, Yue Wu, and Prem Natarajan. 2018.
\newblock \href {https://doi.org/10.1145/3240508.3240707} {Deep multimodal
  image-repurposing detection}.
\newblock In \emph{2018 {ACM} Multimedia Conference on Multimedia Conference,
  {MM} 2018, Seoul, Republic of Korea, October 22-26, 2018}, pages 1337--1345.
  {ACM}.

\bibitem[{Sabir et~al.(2019)Sabir, Cheng, Jaiswal, AbdAlmageed, Masi, and
  Natarajan}]{DBLP:conf/cvpr/SabirCJAMN19}
Ekraam Sabir, Jiaxin Cheng, Ayush Jaiswal, Wael AbdAlmageed, Iacopo Masi, and
  Prem Natarajan. 2019.
\newblock \href
  {http://openaccess.thecvf.com/content\_CVPRW\_2019/html/Media\_Forensics/Sabir\_Recurrent\_Convolutional\_Strategies\_for\_Face\_Manipulation\_Detection\_in\_Videos\_CVPRW\_2019\_paper.html}
  {Recurrent convolutional strategies for face manipulation detection in
  videos}.
\newblock In \emph{{IEEE} Conference on Computer Vision and Pattern Recognition
  Workshops, {CVPR} Workshops 2019, Long Beach, CA, USA, June 16-20, 2019},
  pages 80--87. Computer Vision Foundation / {IEEE}.

\bibitem[{Schlichtkrull et~al.(2023)Schlichtkrull, Guo, and
  Vlachos}]{schlichtkrullGV2023}
Michael~Sejr Schlichtkrull, Zhijiang Guo, and Andreas Vlachos. 2023.
\newblock \href {https://doi.org/10.48550/arXiv.2305.13117} {Averitec: {A}
  dataset for real-world claim verification with evidence from the web}.
\newblock \emph{CoRR}, abs/2305.13117.

\bibitem[{Shahi and Nandini(2020)}]{Shahi2020FakeCovidA}
Gautam~Kishore Shahi and Durgesh Nandini. 2020.
\newblock \href {https://doi.org/10.36190/2020.14} {Fakecovid- {A} multilingual
  cross-domain fact check news dataset for {COVID-19}}.
\newblock In \emph{Workshop Proceedings of the 14th International {AAAI}
  Conference on Web and Social Media, {ICWSM} 2020 Workshops, Atlanta, Georgia,
  {USA} [virtual], June 8, 2020}.

\bibitem[{Shang et~al.(2021)Shang, Kou, Zhang, and Wang}]{ShangKZ021}
Lanyu Shang, Ziyi Kou, Yang Zhang, and Dong Wang. 2021.
\newblock \href {https://doi.org/10.1109/BigData52589.2021.9671928} {A
  multimodal misinformation detector for {COVID-19} short videos on tiktok}.
\newblock In \emph{2021 {IEEE} International Conference on Big Data (Big Data),
  Orlando, FL, USA, December 15-18, 2021}, pages 899--908. {IEEE}.

\bibitem[{Shang et~al.(2022)Shang, Kou, Zhang, and Wang}]{ShangKZW22}
Lanyu Shang, Ziyi Kou, Yang Zhang, and Dong Wang. 2022.
\newblock \href {https://doi.org/10.1145/3485447.3512257} {A duo-generative
  approach to explainable multimodal {COVID-19} misinformation detection}.
\newblock In \emph{{WWW} '22: The {ACM} Web Conference 2022, Virtual Event,
  Lyon, France, April 25 - 29, 2022}, pages 3623--3631. {ACM}.

\bibitem[{Shao et~al.(2023)Shao, Wu, and Liu}]{Shao_2023_CVPR}
Rui Shao, Tianxing Wu, and Ziwei Liu. 2023.
\newblock Detecting and grounding multi-modal media manipulation.
\newblock In \emph{Proceedings of the IEEE/CVF Conference on Computer Vision
  and Pattern Recognition (CVPR)}, pages 6904--6913.

\bibitem[{Silverman(2013)}]{silverman2013verification}
Craig Silverman. 2013.
\newblock Verification handbook.

\bibitem[{Simonyan and Zisserman(2015)}]{SimonyanZ14a}
Karen Simonyan and Andrew Zisserman. 2015.
\newblock \href {http://arxiv.org/abs/1409.1556} {Very deep convolutional
  networks for large-scale image recognition}.
\newblock In \emph{3rd International Conference on Learning Representations,
  {ICLR} 2015, San Diego, CA, USA, May 7-9, 2015, Conference Track
  Proceedings}.

\bibitem[{Singhal et~al.(2022)Singhal, Shah, and
  Kumaraguru}]{DBLP:conf/icwsm/SinghalSK22}
Shivangi Singhal, Rajiv~Ratn Shah, and Ponnurangam Kumaraguru. 2022.
\newblock \href {https://ojs.aaai.org/index.php/ICWSM/article/view/19384}
  {Factdrill: {A} data repository of fact-checked social media content to study
  fake news incidents in india}.
\newblock In \emph{Proceedings of the Sixteenth International {AAAI} Conference
  on Web and Social Media, {ICWSM} 2022, Atlanta, Georgia, USA, June 6-9,
  2022}, pages 1322--1331. {AAAI} Press.

\bibitem[{Sun et~al.(2023)Sun, Qian, Li, and Zhu}]{SunQLZ23}
Tiening Sun, Zhong Qian, Peifeng Li, and Qiaoming Zhu. 2023.
\newblock \href {https://doi.org/10.1145/3591106.3592250} {Graph interactive
  network with adaptive gradient for multi-modal rumor detection}.
\newblock In \emph{Proceedings of the 2023 {ACM} International Conference on
  Multimedia Retrieval, {ICMR} 2023, Thessaloniki, Greece, June 12-15, 2023},
  pages 316--324. {ACM}.

\bibitem[{Suryavardan et~al.(2023{\natexlab{a}})Suryavardan, Mishra,
  Chakraborty, Patwa, Rani, Chadha, Reganti, Das, Sheth, Chinnakotla, Ekbal,
  and Kumar}]{Facitfy23}
S.~Suryavardan, Shreyash Mishra, Megha Chakraborty, Parth Patwa, Anku Rani,
  Aman Chadha, Aishwarya Reganti, Amitava Das, Amit~P. Sheth, Manoj
  Chinnakotla, Asif Ekbal, and Srijan Kumar. 2023{\natexlab{a}}.
\newblock \href {https://doi.org/10.48550/arXiv.2307.10475} {Findings of
  factify 2: Multimodal fake news detection}.
\newblock \emph{CoRR}, abs/2307.10475.

\bibitem[{Suryavardan et~al.(2023{\natexlab{b}})Suryavardan, Mishra, Patwa,
  Chakraborty, Rani, Reganti, Chadha, Das, Sheth, Chinnakotla, Ekbal, and
  Kumar}]{factify2}
S~Suryavardan, Shreyash Mishra, Parth Patwa, Megha Chakraborty, Anku Rani,
  Aishwarya Reganti, Aman Chadha, Amitava Das, Amit~P. Sheth, Manoj
  Chinnakotla, Asif Ekbal, and Srijan Kumar. 2023{\natexlab{b}}.
\newblock \href {https://doi.org/10.48550/arXiv.2304.03897} {Factify 2: {A}
  multimodal fake news and satire news dataset}.
\newblock \emph{CoRR}, abs/2304.03897.

\bibitem[{Szegedy et~al.(2015)Szegedy, Liu, Jia, Sermanet, Reed, Anguelov,
  Erhan, Vanhoucke, and Rabinovich}]{SzegedyWYSRADVR15}
Christian Szegedy, Wei Liu, Yangqing Jia, Pierre Sermanet, Scott Reed, Dragomir
  Anguelov, Dumitru Erhan, Vincent Vanhoucke, and Andrew Rabinovich. 2015.
\newblock \href {https://doi.org/10.1109/CVPR.2015.7298594} {Going deeper with
  convolutions}.
\newblock In \emph{2015 IEEE Conference on Computer Vision and Pattern
  Recognition (CVPR)}, pages 1--9.

\bibitem[{Tan et~al.(2020)Tan, Plummer, and Saenko}]{TanPS20}
Reuben Tan, Bryan~A. Plummer, and Kate Saenko. 2020.
\newblock \href {https://doi.org/10.18653/v1/2020.emnlp-main.163} {Detecting
  cross-modal inconsistency to defend against neural fake news}.
\newblock In \emph{Proceedings of the 2020 Conference on Empirical Methods in
  Natural Language Processing, {EMNLP} 2020, Online, November 16-20, 2020},
  pages 2081--2106. Association for Computational Linguistics.

\bibitem[{Tanwar and Sharma(2020)}]{9182398}
Vidhu Tanwar and Kapil Sharma. 2020.
\newblock \href {https://doi.org/10.1109/ICCSP48568.2020.9182398} {Multi-model
  fake news detection based on concatenation of visual latent features}.
\newblock In \emph{2020 International Conference on Communication and Signal
  Processing (ICCSP)}, pages 1344--1348.

\bibitem[{Thorne and Vlachos(2018)}]{thorne-vlachos-2018-automated}
James Thorne and Andreas Vlachos. 2018.
\newblock \href {https://aclanthology.org/C18-1283} {Automated fact checking:
  Task formulations, methods and future directions}.
\newblock In \emph{Proceedings of the 27th International Conference on
  Computational Linguistics}, pages 3346--3359, Santa Fe, New Mexico, USA.
  Association for Computational Linguistics.

\bibitem[{Thorne et~al.(2018)Thorne, Vlachos, Cocarascu, Christodoulopoulos,
  and Mittal}]{thorne-etal-2018-fact}
James Thorne, Andreas Vlachos, Oana Cocarascu, Christos Christodoulopoulos, and
  Arpit Mittal. 2018.
\newblock \href {https://doi.org/10.18653/v1/W18-5501} {The fact extraction and
  {VER}ification ({FEVER}) shared task}.
\newblock In \emph{Proceedings of the First Workshop on Fact Extraction and
  {VER}ification ({FEVER})}, pages 1--9, Brussels, Belgium. Association for
  Computational Linguistics.

\bibitem[{Vaswani et~al.(2017)Vaswani, Shazeer, Parmar, Uszkoreit, Jones,
  Gomez, Kaiser, and Polosukhin}]{VaswaniSPUJGKP17}
Ashish Vaswani, Noam Shazeer, Niki Parmar, Jakob Uszkoreit, Llion Jones,
  Aidan~N. Gomez, Lukasz Kaiser, and Illia Polosukhin. 2017.
\newblock \href
  {https://proceedings.neurips.cc/paper/2017/hash/3f5ee243547dee91fbd053c1c4a845aa-Abstract.html}
  {Attention is all you need}.
\newblock In \emph{Advances in Neural Information Processing Systems 30: Annual
  Conference on Neural Information Processing Systems 2017, December 4-9, 2017,
  Long Beach, CA, {USA}}, pages 5998--6008.

\bibitem[{Wang et~al.(2022)Wang, Wu, Ouyang, Han, Chen, Jiang, and
  Li}]{DBLP:conf/mir/WangWOHCJL22}
Junke Wang, Zuxuan Wu, Wenhao Ouyang, Xintong Han, Jingjing Chen, Yu{-}Gang
  Jiang, and Ser{-}Nam Li. 2022.
\newblock \href {https://doi.org/10.1145/3512527.3531415} {{M2TR:} multi-modal
  multi-scale transformers for deepfake detection}.
\newblock In \emph{{ICMR} '22: International Conference on Multimedia
  Retrieval, Newark, NJ, USA, June 27 - 30, 2022}, pages 615--623. {ACM}.

\bibitem[{Wang et~al.(2020)Wang, Juefei{-}Xu, Huang, Guo, Xie, Ma, and
  Liu}]{WangJHGXML20}
Run Wang, Felix Juefei{-}Xu, Yihao Huang, Qing Guo, Xiaofei Xie, Lei Ma, and
  Yang Liu. 2020.
\newblock \href {https://doi.org/10.1145/3394171.3413716} {Deepsonar: Towards
  effective and robust detection of ai-synthesized fake voices}.
\newblock In \emph{{MM} '20: The 28th {ACM} International Conference on
  Multimedia, Virtual Event / Seattle, WA, USA, October 12-16, 2020}, pages
  1207--1216. {ACM}.

\bibitem[{Wang et~al.(2018)Wang, Ma, Jin, Yuan, Xun, Jha, Su, and
  Gao}]{WangMJYXJSG18}
Yaqing Wang, Fenglong Ma, Zhiwei Jin, Ye~Yuan, Guangxu Xun, Kishlay Jha, Lu~Su,
  and Jing Gao. 2018.
\newblock \href {https://doi.org/10.1145/3219819.3219903} {{EANN:} event
  adversarial neural networks for multi-modal fake news detection}.
\newblock In \emph{Proceedings of the 24th {ACM} {SIGKDD} International
  Conference on Knowledge Discovery {\&} Data Mining, {KDD} 2018, London, UK,
  August 19-23, 2018}, pages 849--857. {ACM}.

\bibitem[{Wodajo and Atnafu(2021)}]{DBLP:journals/corr/abs-2102-11126}
Deressa Wodajo and Solomon Atnafu. 2021.
\newblock \href {http://arxiv.org/abs/2102.11126} {Deepfake video detection
  using convolutional vision transformer}.
\newblock \emph{CoRR}, abs/2102.11126.

\bibitem[{Wu et~al.(2023{\natexlab{a}})Wu, Zhou, and
  Zhang}]{wu-etal-generalizable}
Haiwei Wu, Jiantao Zhou, and Shile Zhang. 2023{\natexlab{a}}.
\newblock \href {https://doi.org/10.48550/arXiv.2305.13800} {Generalizable
  synthetic image detection via language-guided contrastive learning}.
\newblock \emph{CoRR}, abs/2305.13800.

\bibitem[{Wu et~al.(2023{\natexlab{b}})Wu, Liao, and Ou}]{WuLO2023}
Xiaoshuai Wu, Xin Liao, and Bo~Ou. 2023{\natexlab{b}}.
\newblock \href {https://doi.org/10.48550/arXiv.2305.06321} {Sepmark: Deep
  separable watermarking for unified source tracing and deepfake detection}.
\newblock \emph{CoRR}, abs/2305.06321.

\bibitem[{Wu et~al.(2021)Wu, Zhan, Zhang, Wang, and
  Xu}]{wu-etal-2021-multimodal}
Yang Wu, Pengwei Zhan, Yunjian Zhang, Liming Wang, and Zhen Xu. 2021.
\newblock \href {https://doi.org/10.18653/v1/2021.findings-acl.226} {Multimodal
  fusion with co-attention networks for fake news detection}.
\newblock In \emph{Findings of the Association for Computational Linguistics:
  ACL-IJCNLP 2021}, pages 2560--2569, Online. Association for Computational
  Linguistics.

\bibitem[{Wu et~al.(2019)Wu, AbdAlmageed, and
  Natarajan}]{DBLP:conf/cvpr/0001AN19}
Yue Wu, Wael AbdAlmageed, and Premkumar Natarajan. 2019.
\newblock \href {https://doi.org/10.1109/CVPR.2019.00977} {Mantra-net:
  Manipulation tracing network for detection and localization of image
  forgeries with anomalous features}.
\newblock In \emph{{IEEE} Conference on Computer Vision and Pattern
  Recognition, {CVPR} 2019, Long Beach, CA, USA, June 16-20, 2019}, pages
  9543--9552. Computer Vision Foundation / {IEEE}.

\bibitem[{Wu et~al.(2015)Wu, Kinnunen, Evans, Yamagishi, Hanil{\c{c}}i,
  Sahidullah, and Sizov}]{WuKEYHSS15}
Zhizheng Wu, Tomi Kinnunen, Nicholas W.~D. Evans, Junichi Yamagishi, Cemal
  Hanil{\c{c}}i, Md. Sahidullah, and Aleksandr Sizov. 2015.
\newblock \href
  {http://www.isca-speech.org/archive/interspeech\_2015/i15\_2037.html}
  {Asvspoof 2015: the first automatic speaker verification spoofing and
  countermeasures challenge}.
\newblock In \emph{{INTERSPEECH} 2015, 16th Annual Conference of the
  International Speech Communication Association, Dresden, Germany, September
  6-10, 2015}, pages 2037--2041. {ISCA}.

\bibitem[{Yao et~al.(2022)Yao, Shah, Sun, Cho, and Huang}]{Yao2022}
Barry~Menglong Yao, Aditya Shah, Lichao Sun, Jin{-}Hee Cho, and Lifu Huang.
  2022.
\newblock \href {https://doi.org/10.48550/arXiv.2205.12487} {End-to-end
  multimodal fact-checking and explanation generation: {A} challenging dataset
  and models}.
\newblock \emph{CoRR}, abs/2205.12487.

\bibitem[{Yi et~al.(2021)Yi, Bai, Tao, Ma, Tian, Wang, Wang, and
  Fu}]{YiBTMTWWF21}
Jiangyan Yi, Ye~Bai, Jianhua Tao, Haoxin Ma, Zhengkun Tian, Chenglong Wang, Tao
  Wang, and Ruibo Fu. 2021.
\newblock \href {https://doi.org/10.21437/Interspeech.2021-930} {Half-truth:
  {A} partially fake audio detection dataset}.
\newblock In \emph{Interspeech 2021, 22nd Annual Conference of the
  International Speech Communication Association, Brno, Czechia, 30 August - 3
  September 2021}, pages 1654--1658. {ISCA}.

\bibitem[{Zakharov et~al.(2019)Zakharov, Shysheya, Burkov, and
  Lempitsky}]{ZakharovSBL19}
Egor Zakharov, Aliaksandra Shysheya, Egor Burkov, and Victor~S. Lempitsky.
  2019.
\newblock \href {https://doi.org/10.1109/ICCV.2019.00955} {Few-shot adversarial
  learning of realistic neural talking head models}.
\newblock In \emph{2019 {IEEE/CVF} International Conference on Computer Vision,
  {ICCV} 2019, Seoul, Korea (South), October 27 - November 2, 2019}, pages
  9458--9467. {IEEE}.

\bibitem[{Zampoglou et~al.(2015)Zampoglou, Papadopoulos, and
  Kompatsiaris}]{ZampoglouPK15}
Markos Zampoglou, Symeon Papadopoulos, and Yiannis Kompatsiaris. 2015.
\newblock \href {https://doi.org/10.1109/ICMEW.2015.7169839} {Detecting image
  splicing in the wild {(WEB)}}.
\newblock In \emph{2015 {IEEE} International Conference on Multimedia {\&} Expo
  Workshops, {ICME} Workshops 2015, Turin, Italy, June 29 - July 3, 2015},
  pages 1--6. {IEEE} Computer Society.

\bibitem[{Zeng et~al.(2021)Zeng, Abumansour, and
  Zubiaga}]{DBLP:journals/llc/ZengAZ21}
Xia Zeng, Amani~S. Abumansour, and Arkaitz Zubiaga. 2021.
\newblock \href {https://doi.org/10.1111/lnc3.12438} {Automated fact-checking:
  {A} survey}.
\newblock \emph{Lang. Linguistics Compass}, 15(10).

\bibitem[{Zhang et~al.(2018)Zhang, Shang, Geng, Lai, Li, Zhu, Amin, and
  Wang}]{ZhangSGLLZA018}
Daniel~Yue Zhang, Lanyu Shang, Biao Geng, Shuyue Lai, Ke~Li, Hongmin Zhu, Md.
  Tanvir~Al Amin, and Dong Wang. 2018.
\newblock \href {https://doi.org/10.1109/BigData.2018.8622344} {Fauxbuster: {A}
  content-free fauxtography detector using social media comments}.
\newblock In \emph{{IEEE} International Conference on Big Data {(IEEE} BigData
  2018), Seattle, WA, USA, December 10-13, 2018}, pages 891--900. {IEEE}.

\bibitem[{Zhang et~al.(2023)Zhang, Trinh, Cao, Cui, and Liu}]{zhang-etal-2023}
Yizhou Zhang, Loc Trinh, Defu Cao, Zijun Cui, and Yan Liu. 2023.
\newblock \href {https://doi.org/10.48550/arXiv.2304.07633} {Detecting
  out-of-context multimodal misinformation with interpretable neural-symbolic
  model}.
\newblock \emph{CoRR}, abs/2304.07633.

\bibitem[{Zheng et~al.(2022)Zheng, Zhang, Guo, Wang, Zang, and
  Zhang}]{ZhengZGWZ022}
Jiaqi Zheng, Xi~Zhang, Sanchuan Guo, Quan Wang, Wenyu Zang, and Yongdong Zhang.
  2022.
\newblock \href {https://doi.org/10.24963/ijcai.2022/335} {{MFAN:} multi-modal
  feature-enhanced attention networks for rumor detection}.
\newblock In \emph{Proceedings of the Thirty-First International Joint
  Conference on Artificial Intelligence, {IJCAI} 2022, Vienna, Austria, 23-29
  July 2022}, pages 2413--2419. ijcai.org.

\bibitem[{Zheng et~al.(2021)Zheng, Bao, Chen, Zeng, and
  Wen}]{DBLP:conf/iccv/ZhengB0ZW21}
Yinglin Zheng, Jianmin Bao, Dong Chen, Ming Zeng, and Fang Wen. 2021.
\newblock \href {https://doi.org/10.1109/ICCV48922.2021.01477} {Exploring
  temporal coherence for more general video face forgery detection}.
\newblock In \emph{2021 {IEEE/CVF} International Conference on Computer Vision,
  {ICCV} 2021, Montreal, QC, Canada, October 10-17, 2021}, pages 15024--15034.
  {IEEE}.

\bibitem[{Zhou et~al.(2018)Zhou, Han, Morariu, and Davis}]{Zhou_2018_CVPR}
Peng Zhou, Xintong Han, Vlad~I. Morariu, and Larry~S. Davis. 2018.
\newblock Learning rich features for image manipulation detection.
\newblock In \emph{Proceedings of the IEEE Conference on Computer Vision and
  Pattern Recognition (CVPR)}.

\bibitem[{Zlatkova et~al.(2019)Zlatkova, Nakov, and
  Koychev}]{zlatkova-etal-2019-fact}
Dimitrina Zlatkova, Preslav Nakov, and Ivan Koychev. 2019.
\newblock \href {https://doi.org/10.18653/v1/D19-1216} {Fact-checking meets
  fauxtography: Verifying claims about images}.
\newblock In \emph{Proceedings of the 2019 Conference on Empirical Methods in
  Natural Language Processing and the 9th International Joint Conference on
  Natural Language Processing (EMNLP-IJCNLP)}, pages 2099--2108, Hong Kong,
  China. Association for Computational Linguistics.

\bibitem[{Zong et~al.(2023)Zong, Zhang, Shang, and Wang}]{ZongZSW23}
Ruohan Zong, Yang Zhang, Lanyu Shang, and Dong Wang. 2023.
\newblock \href {https://doi.org/10.1145/3543507.3583869} {Contrastfaux: Sparse
  semi-supervised fauxtography detection on the web using multi-view
  contrastive learning}.
\newblock In \emph{Proceedings of the {ACM} Web Conference 2023, {WWW} 2023,
  Austin, TX, USA, 30 April 2023 - 4 May 2023}, pages 3994--4003. {ACM}.

\end{thebibliography}
\bibliographystyle{acl_natbib}

\appendix

\section{Methodology}
\label{sec:methodology}

We applied the following methodological approach to find and select relevant research papers for the survey. 

First, after defining the research scope, we collected pivotal, highly-cited work (e.g. \citet{NakamuraLW20}) and related surveys (e.g. \cite{alam-etal-2022-survey}), resulting in $25$ papers, as well as papers citing or cited by these works.
We collected further works using the scholarly search engines Google Scholar\footnote{\url{https://scholar.google.com/}}, Semantic Scholar\footnote{\url{https://www.semanticscholar.org/}}, DBLP\footnote{\url{https://dblp.org/}} and ACL Anthology\footnote{\url{https://aclanthology.org/}}, and keyword-based search with Cartesian products of following keyword sets: \{``fact checking'', ``fact verification'', ``misinformation'', ``disinformation'', ``fake news''\}, \{``multimodal'', ``text'', ``image'', ``audio'', ``video''\}, and \{``machine learning'', ``automated''\}. The databases were queried primarily during the time frame July $26, 2022$ and August $10, 2022$. 
This step resulted in a collection of $123$ papers.

We manually screened and filtered the papers based on abstracts and introduction sections, before creating an overview of papers across the following dimensions: $(1)$ modality; $(2)$ fact-checking task; $(3)$ contribution type (i.e. dataset, modeling approach, demo); $(4)$ paper type (i.e. survey, position paper, solution paper (e.g. introducing a new benchmark or modeling approach), or evaluation paper (e.g. investigating previously proposed approaches)). Papers were mostly excluded because they focused on other tasks than fact-checking (e.g. hate speech detection) or used modalities out of our scope (e.g. tables). Moreover, during the screening process we found and added further related works, and concluded the screening with $84$ unique papers. 

The taxonomy of tasks (Section \ref{sec:taskformulation}) was created in an iterative manner starting with the task labels we assigned to works during screening. As a starting point we also used taxonomies of text-only fact-checking surveys~\citep{guo-etal-2022-survey, thorne-vlachos-2018-automated} and adapted them for multimodal fact-checking works.


\section{Examples: multimodal misinformation}
\label{sec:examples}

\begin{figure}
\centering
\includegraphics[scale=0.12]{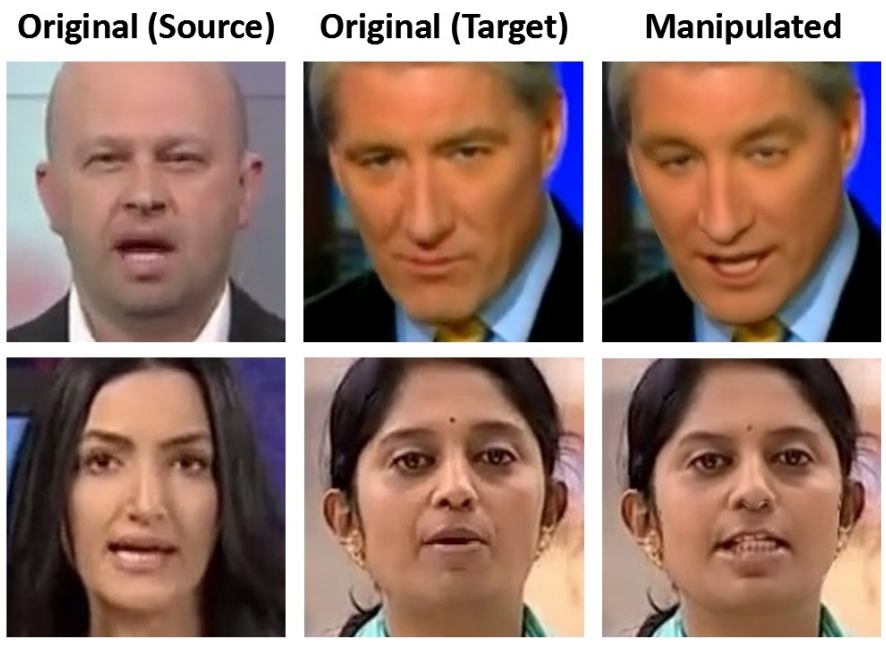}
\caption{\label{fig:example_appendix} Example from the \emph{FaceForensic} video manipulation dataset~\citep{Andreas18} showing the manipulation generation approach. 
}
\end{figure}

\begin{figure}
\centering
\includegraphics[scale=0.18]{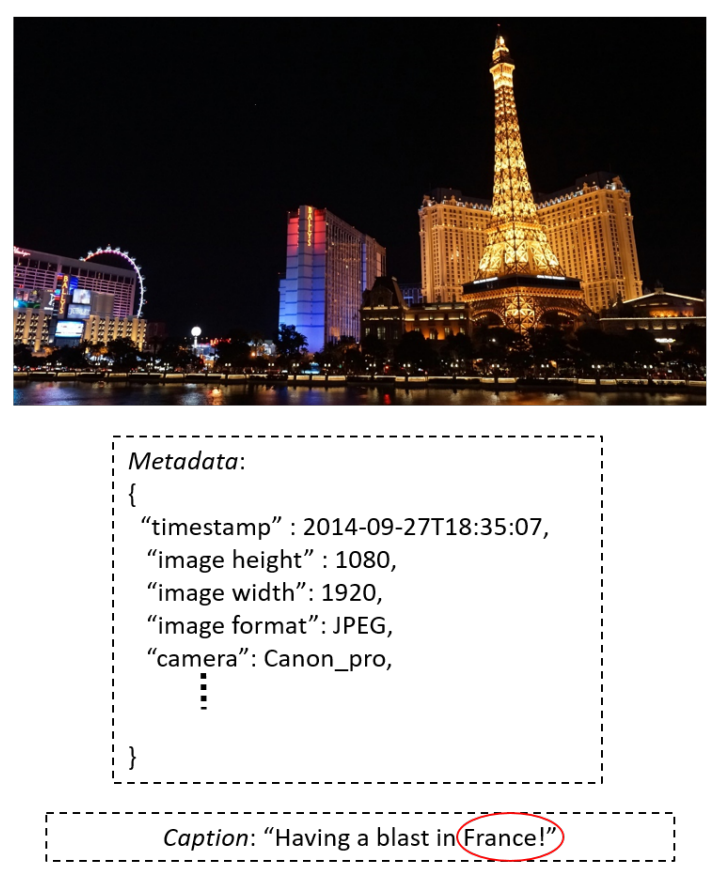}
\caption{\label{fig:example_appendix} An entry from the \emph{MAIM} dataset~\citep{JaiswalSAN17} showing an image/text claim with metadata. 
}
\end{figure}

\begin{figure}
\centering
\includegraphics[scale=0.22]{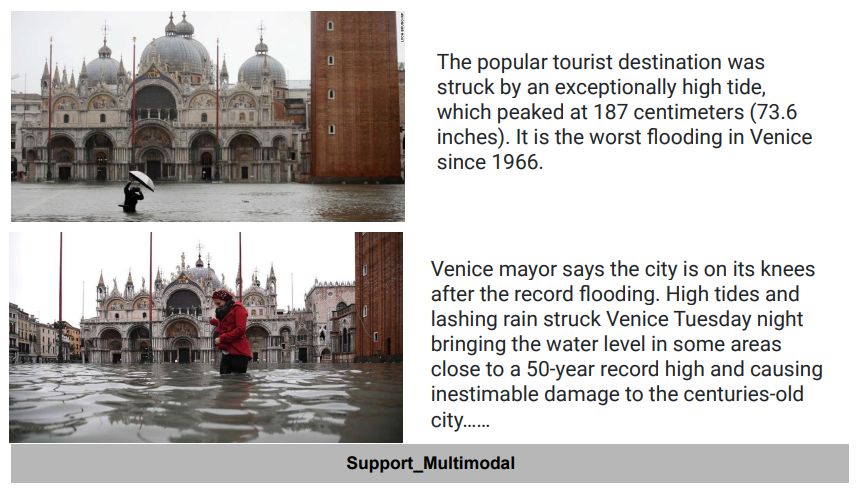}
\caption{\label{fig:example_appendix} An entry from the \emph{Factify} dataset~\citep{factify2} depicting an image/text claim and supporting image/text evidence document. 
}
\end{figure}

\begin{figure}
\centering
\includegraphics[scale=0.14]{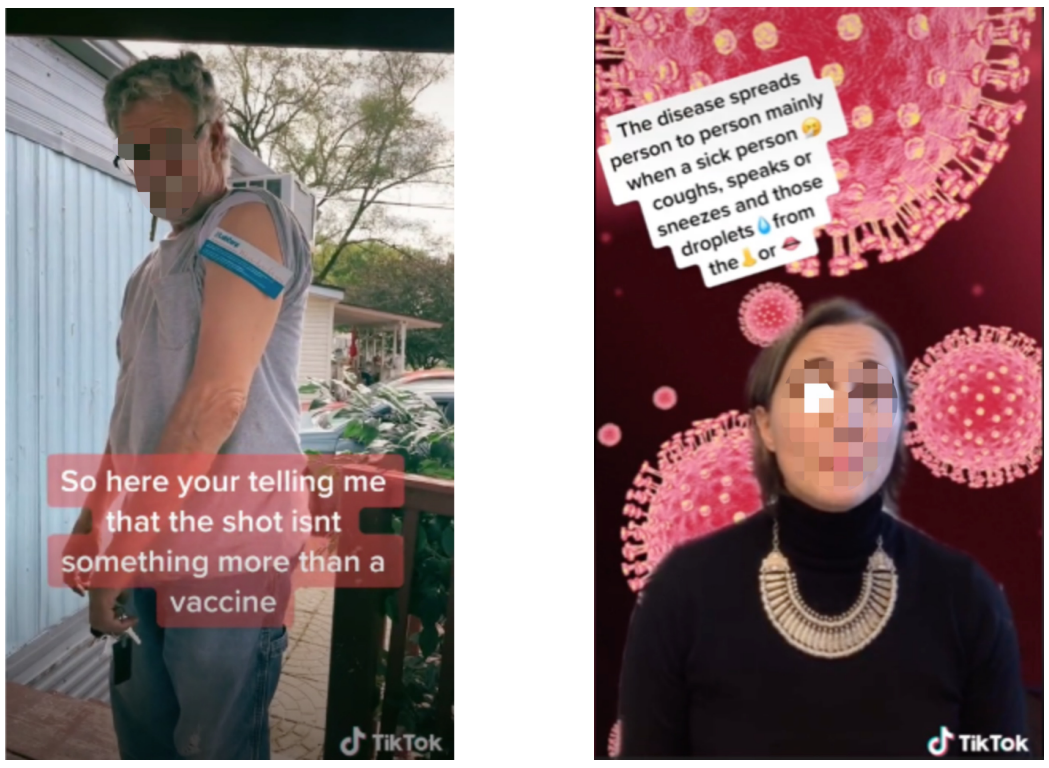}
\caption{\label{fig:example_appendix} Left a misleading, right a non-misleading video screenshot from the \citet{ShangKZ021} dataset on COVID-19 TikTok Short Videos. 
}
\end{figure}



\end{document}